\documentclass[letterpaper]{article} 
\usepackage{aaai2026}  
\usepackage{times}  
\usepackage{helvet}  
\usepackage{courier}  
\usepackage[hyphens]{url}  
\usepackage{graphicx} 
\usepackage{natbib}  
\usepackage{caption} 
\usepackage{algorithm}
\usepackage{algorithmic}
\usepackage{multirow}
\usepackage{comment}

\usepackage[table,xcdraw]{xcolor}
\usepackage{pifont}
\usepackage{booktabs}
\usepackage{array}
\usepackage{arydshln}
\usepackage{amsfonts}
\usepackage{makecell}
\usepackage{amsmath}

\usepackage{newfloat}
\usepackage{listings}
\DeclareCaptionStyle{ruled}{labelfont=normalfont,labelsep=colon,strut=off} 
\floatstyle{ruled}
\newfloat{listing}{tb}{lst}{}
\floatname{listing}{Listing}
\title{STRIDE-QA: Visual Question Answering Dataset for \\Spatiotemporal Reasoning in Urban Driving Scenes}

\author{
    Keishi Ishihara\textsuperscript{\rm 1}\equalcontrib,\
    Kento Sasaki\textsuperscript{\rm 1, 2}\equalcontrib,\
    Tsubasa Takahashi\textsuperscript{\rm 1},\
    Daiki Shiono\textsuperscript{\rm 1, 3},\
    Yu Yamaguchi\textsuperscript{\rm 1}
}
\affiliations{
    \textsuperscript{\rm 1}Turing Inc.\quad \\
    \textsuperscript{\rm 2}University of Tsukuba\quad \\
    \textsuperscript{\rm 3}Tohoku University \\
    \{keishi.ishihara, kento.sasaki\}@turing-motors.com
}

\begin{document}

\maketitle
\renewcommand\twocolumn[1][]{#1}

\begin{abstract}
Vision-Language Models (VLMs) have been applied to autonomous driving to support decision-making in complex real-world scenarios. However, their training on static, web-sourced image-text pairs fundamentally limits the precise spatiotemporal reasoning required to understand and predict dynamic traffic scenes. We address this critical gap with STRIDE-QA, a large-scale visual question answering (VQA) dataset for physically grounded reasoning from an ego-centric perspective. Constructed from 100 hours of multi-sensor driving data in Tokyo, capturing diverse and challenging conditions, STRIDE-QA is the largest VQA dataset for spatiotemporal reasoning in urban driving, offering 16\,M QA pairs over 270\,K frames. Grounded by dense, automatically generated annotations including 3D bounding boxes, segmentation masks, and multi-object tracks, the dataset uniquely supports both object-centric and ego-centric reasoning through three novel QA tasks that require spatial localization and temporal prediction. Our benchmarks demonstrate that existing VLMs struggle significantly, with near-zero scores on prediction consistency. In contrast, VLMs fine-tuned on STRIDE-QA exhibit dramatic performance gains, achieving 55\% success in spatial localization and 28\% consistency in future motion prediction, compared to near-zero scores from general-purpose VLMs. Therefore, STRIDE-QA establishes a comprehensive foundation for developing more reliable VLMs for safety-critical autonomous systems.
\end{abstract}

\begin{links}
\link{Dataset}{https://turingmotors.github.io/stride-qa/}
\link{Extended version}{https://arxiv.org/abs/2508.10427}
\end{links}

\begin{figure*}[t]
  \centering
  \includegraphics[width=0.99\linewidth]{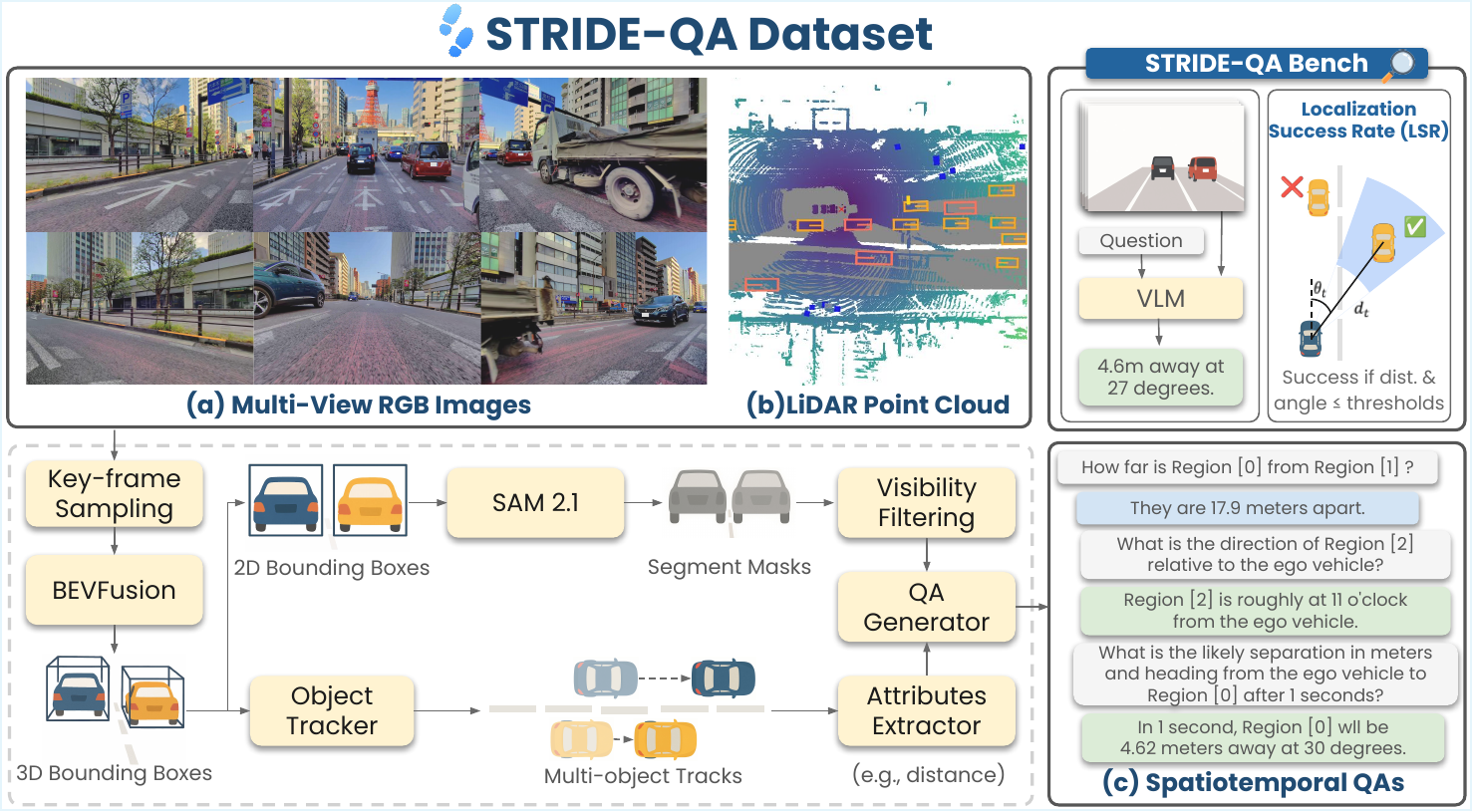}
  \caption{STRIDE-QA is a large-scale VQA dataset for spatiotemporal reasoning in autonomous driving, comprising 270\,K frames and 16\,M QA pairs from over 100 hours of urban driving in Tokyo. (a) It includes multi-view RGB images and (b) LiDAR point clouds, processed via a modular pipeline with 3D object detection, segmentation, tracking, and visibility filtering to produce spatially and temporally grounded annotations. (c) The annotations enable object-centric, ego-centric spatial, and spatiotemporal QA tasks, allowing structured evaluation of physically grounded reasoning over time.}
  \label{fig:overview}
\end{figure*}

\section{Introduction}
\label{sec:intro}

Recent advances in Vision-Language Models (VLMs) have led to remarkable progress across multimodal tasks such as image captioning and visual question answering~\cite{llava, qwen2_5}. These models, trained on large-scale image-text datasets, exhibit strong semantic understanding and generalization capabilities. Motivated by this success, VLMs have been applied to Physical AI~\cite{nvidia2025cosmos-reason1}, such as in robotics~\cite{rt2, black2024pi0visionlanguageactionflowmodel} and autonomous driving~\cite{DriveLM, lingoqa}.

These approaches underscore the promise of VLMs in achieving holistic scene understanding and high-level driving decision-making. However, this promise is fundamentally limited by the nature of their training data: most VLMs are trained on static, web-sourced image-text pairs and consequently lack the spatial reasoning capabilities essential for real-world applications~\cite{blink}. This limitation is particularly critical in autonomous driving, where the lack of appropriate training data remains a significant challenge.

To address this limitation, we introduce \textbf{STRIDE-QA} (\textbf{S}patio\textbf{T}emporal \textbf{R}easoning \textbf{I}n \textbf{D}riving Scenarios for \textbf{E}go-centric Visual \textbf{Q}uestion \textbf{A}nswering), a large-scale VQA dataset designed for fine-grained spatial and spatiotemporal reasoning in real-world driving scenes. An overview of the dataset and its automated annotation pipeline is shown in Figure 1. The dataset is constructed from over 100 hours of multi-sensor driving data collected in Tokyo, capturing diverse and challenging scenarios including traffic congestion, construction zones, and pedestrian-dense intersections. It contains over 16\,M QA pairs generated through a fully modular and scalable annotation pipeline that integrates 3D object detection, multi-object tracking, and instance segmentation.

STRIDE-QA is specifically constructed to enable supervised training and evaluation on three core reasoning tasks:

\begin{itemize}
\item \textbf{Object-centric Spatial QA}: Assessing spatial relations between non-ego agents (vehicles, pedestrians, etc.).
\item \textbf{Ego-centric Spatial QA}: Describing agents' distance, orientation, and size relative to the ego vehicle.
\item \textbf{Ego-centric Spatiotemporal QA}: Predicting how agent–ego spatial relations evolve over time.
\end{itemize}

These tasks are designed to systematically measure the spatial and predictive reasoning capabilities essential for downstream planning and decision making in safety-critical urban environments. By grounding each QA pair in physically and temporally consistent annotations, STRIDE-QA provides a comprehensive foundation for training and benchmarking VLMs in real-world autonomous driving.

\textbf{Contributions}
The main contributions of this paper are summarized as follows:
\begin{itemize}
    \item We define three novel ego-centric VQA tasks that jointly require spatial grounding and short-term predictive reasoning, addressing core challenges autonomous driving systems face in complex traffic scenes.
    \item We present STRIDE-QA, a large-scale dataset containing 16\,M QA pairs densely annotated over 270\,K video frames from urban driving, which enables supervised training of VLMs on fine-grained spatial and short-term temporal reasoning grounded in real-world traffic dynamics.
    \item We demonstrate that existing general-purpose VLMs struggle with spatiotemporal reasoning, while models fine-tuned on STRIDE-QA significantly outperform baselines. Our best model, STRIDE-Qwen2.5-VL-7B, achieves state-of-the-art performance, highlighting the effectiveness of our dataset for spatiotemporal understanding in driving scenes.
\end{itemize}

These contributions represent a key step toward integrating VLMs into real-world autonomous systems. By shifting from general-purpose vision-language understanding to physically grounded, ego-centric reasoning, STRIDE-QA bridges the gap between large-scale multimodal pretraining and the demands of Physical AI. It offers not only a benchmark for spatiotemporal VQA but also advances spatial understanding in dynamic scenes, promoting more trustworthy VLMs for autonomous driving.

\begin{table*}[t]
\centering
\scalebox{1.08}{
\scriptsize
\begin{tabular}{lcccccccccc}
\toprule
\textbf{Dataset} & \textbf{Data Source} & \textbf{Modality} & \textbf{\# Video} & \textbf{\# QA} & \multicolumn{2}{c}{\textbf{Viewpoint}} & \multicolumn{4}{c}{\textbf{QA Type}} \\
 & & & & & Obj. & Ego & S-Q & S-N & ST-Q & ST-N \\
\midrule
Spatial-VQA~\cite{chen2024spatialvlm} & Web images & RGB & --- & 200\,M & \ding{51} &  & \ding{51} & \ding{51} &  &  \\
Open Spatial Dataset~\cite{cheng2024spatialrgpt} & OpenImages & RGB & --- & 8.7\,M & \ding{51} &  & \ding{51} & \ding{51} &  &  \\
\midrule
Refer-KITTI~\cite{Wu_2023_CVPR} & KITTI & RGB + LiDAR & 6 h & 818 &  & \ding{51} & \ding{51} &  & \ding{51} &  \\
ToD3Cap~\cite{tod3cap} & nuScenes & RGB + LiDAR & 5.5 h & 468\,K & \ding{51} & \ding{51} & \ding{51} &  &  &  \\
nuScenes-QA~\cite{qian2024nuscenesqa} & nuScenes & RGB + LiDAR & 5.5 h & 460\,K & \ding{51} & \ding{51} & \ding{51} &  & \ding{51} &  \\
NuPrompt~\cite{nuprompt} & nuScenes & RGB + LiDAR & 5.5 h & 87.3\,K &  & \ding{51} & \ding{51} &  & \ding{51} &  \\
nuPlanQA~\cite{nuplanqa} & nuPlan & RGB + LiDAR & 119 h & 1\,M & \ding{51} & \ding{51} & \ding{51} &  & \ding{51} &  \\
TUMTraffic-VideoQA~\cite{zhou2025tumtraf} & AD & Self-collected & RGB & $\leq$33.3 h & 87.3\,K & \ding{51} & \ding{51} & \ding{51} &  & \ding{51} \\
\rowcolor{cyan!10}
\midrule
\textbf{STRIDE-QA (Ours)} & Self-collected & RGB + LiDAR & 100 h & 16M & \ding{51} & \ding{51} & \ding{51} & \ding{51} &  & \ding{51} \\
\bottomrule
\end{tabular}
}
\caption{
Comparison of STRIDE-QA with existing visual question answering datasets by data source, modality, scale, viewpoint, and QA types.
S-Q, S-N, ST-Q, and ST-N denote Spatial Qualitative, Spatial Numerical, Spatiotemporal Qualitative, and Spatiotemporal Numerical.
STRIDE-QA is the first dataset to provide large-scale, ego-centric spatiotemporal supervision.
}
\label{tab:comparison}
\end{table*}

\section{Related Work}
\label{sec:relatedwork}

VLMs have demonstrated strong performance on multimodal tasks by training on 2D vision-language datasets such as VQA v2~\cite{goyal2017making} and GQA~\cite{ainslie2023gqa}. However, these datasets lack 3D spatial information, which limits their applicability to physically grounded reasoning in real-world environments.

To address this gap, spatially-aware VLMs such as SpatialVLM~\cite{chen2024spatialvlm} and Spatial-RGPT~\cite{cheng2024spatialrgpt} incorporate geometric annotations that enable metric reasoning in 3D. Nevertheless, these models remain confined to static, single-frame inputs and cannot capture the temporal dynamics critical for applications such as autonomous driving.

In response, several VQA datasets have been introduced for driving scenarios. For example, nuScenes-QA~\cite{nuplanqa} leverages multi-view videos to formulate spatial queries. ToD3Cap~\cite{tod3cap} and NuPrompt~\cite{nuprompt} focus on ego-centric spatial grounding, whereas Refer-KITTI~\cite{wu2023referring} and TUMTraffic-VideoQA~\cite{zhou2025tumtraf} explore video-based QA and referring expressions.

While these datasets represent significant progress toward incorporating spatial and temporal reasoning into vision-language models, several challenges remain unaddressed. Notably, most existing benchmarks are designed to support either object-centric or ego-centric reasoning, but not both. This distinction is important because object-centric reasoning is crucial for capturing interactions among agents, whereas ego-centric reasoning enables spatial understanding relative to the ego vehicle. In addition, many datasets lack temporally aligned 3D annotations, essential for short-horizon predictive reasoning. Scalability also remains a challenge, as real-world driving scenarios involve high visual diversity and complexity. A detailed comparison of these datasets is provided in Table~\ref{tab:comparison}.

To fill this gap, we introduce STRIDE-QA. It comprises over 100 hours of synchronized LiDAR and multi-view RGB data, and defines three QA tasks: object-centric spatial, ego-centric spatial, and ego-centric spatiotemporal. These tasks collectively support fine-grained, predictive reasoning under complex traffic conditions.

\section{Spatial and Spatiotemporal QA Definition}
\label{sec:preliminary}

This section outlines key requirements for spatiotemporal QA in autonomous driving before introducing our dataset and benchmark. To support fine-grained reasoning in urban traffic scenarios, we define three VQA categories targeting distinct aspects of scene understanding: (1) relationships between surrounding agents, (2) ego-to-agent spatial relations, and (3) short-term prediction of agent dynamics.

\begin{figure*}[t]
    \centering
    \includegraphics[width=0.99\linewidth]{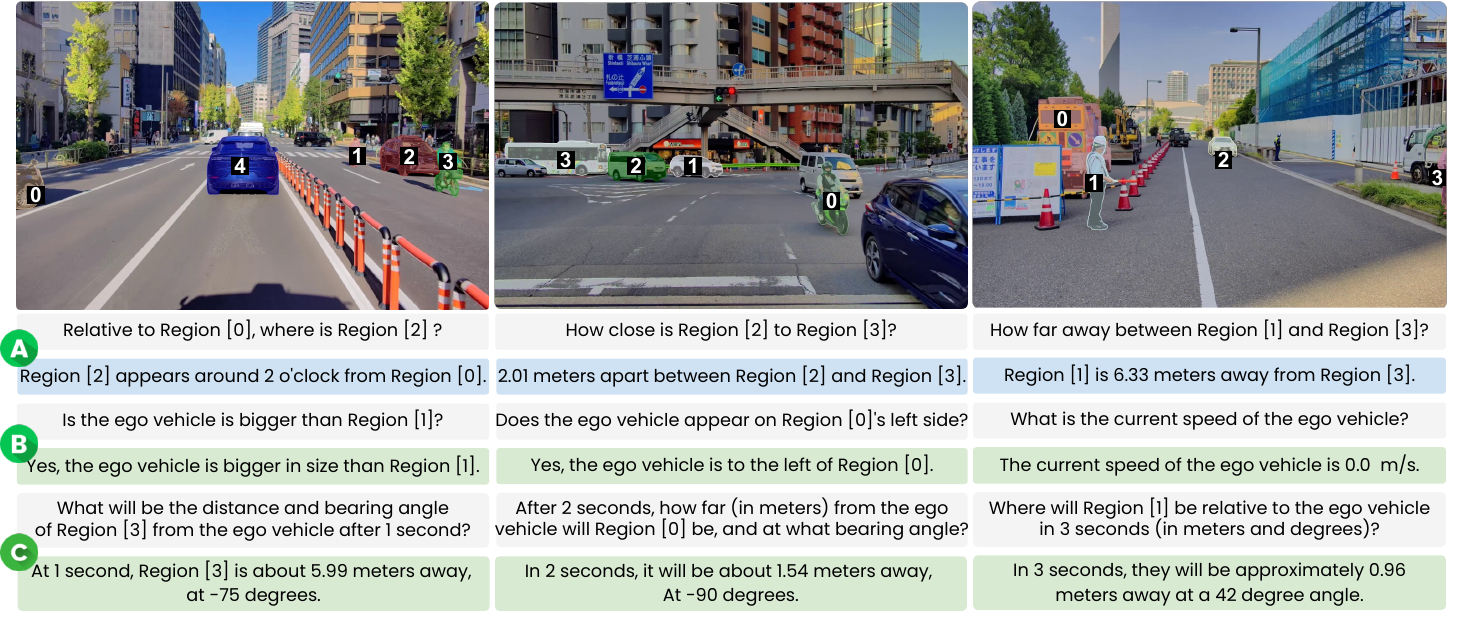}
    \caption{Example data from our STRIDE-QA dataset. From top to bottom, each QA pair corresponds to (A) Object-centric Spatial QA, (B) Ego-centric Spatial QA, and (C) Ego-centric Spatiotemporal QA.}
    \label{fig:data_samples}
\end{figure*}

\noindent\textbf{Object-centric Spatial QA.}\quad
This category targets spatial relationships between two surrounding agents based on a single current frame. It includes both qualitative questions (e.g., relative position or yes/no queries) and quantitative ones that require numerical answers (e.g., ``1.53 m away'' or ``5 degrees''), as illustrated in Figure~\ref{fig:data_samples} (A).

\noindent\textbf{Ego-centric Spatial QA.}\quad
Understanding the scene from the ego-centric perspective is critical for autonomous driving. To support this, we design a category of questions that target the spatial relationship between the ego vehicle and a single surrounding agent, based on a single current frame. We construct both qualitative and quantitative questions involving distance, relative direction, and object size with respect to the ego vehicle, as illustrated in Figure~\ref{fig:data_samples} (B).

\noindent\textbf{Ego-centric Spatiotemporal QA.}\quad
This category extends Ego-centric Spatial QA by incorporating short-term temporal prediction.
The task takes four context frames sampled at 2 Hz as input and aims to forecast the following physical quantities at future time horizons \(t \in \{1,\,2,\,3\}\;\text{s}\):
\begin{itemize}
    \item \textbf{Distance} between the ego vehicle and the target agent
    \item \textbf{Heading angle} from the ego vehicle to the target agent
    \item \textbf{Velocity} of both the ego vehicle and the target agent
\end{itemize}
This task's QA examples are shown in Figure~\ref{fig:data_samples} (C).

\section{STRIDE-QA Dataset}
\label{sec:dataset}

We introduce \textbf{STRIDE-QA} (\textbf{S}patio\textbf{T}emporal \textbf{R}easoning \textbf{i}n \textbf{D}riving Scenarios for \textbf{E}go-centric Visual \textbf{Q}uestion \textbf{A}nswering), a large-scale benchmark designed to advance ego-centric spatiotemporal reasoning in urban driving. STRIDE-QA consists of \textbf{270\,K} frames, \textbf{268\,K} unique object pairs, and \textbf{16\,M} QA pairs.

Section~\ref{subsec:data_collection} describes the data collection setup, and Section~\ref{subsec:pipeline} details the automated annotation pipeline.

\subsection{Driving Data Collection}
\label{subsec:data_collection}

\noindent\textbf{Driving Areas.}\quad
We collected over 100 hours of driving data in Tokyo, which is known for its challenging environment caused by traffic congestion, complex regulations such as no-parking zones and one-way streets, and the high density of pedestrians and cyclists. The city includes downtown districts, residential neighborhoods, and suburban areas.

\noindent\textbf{Sensor Setup.}\quad
We utilize a multi-purpose vehicle equipped with a sensor suite consisting of a 64-channel LiDAR and six cameras, which are synchronized and downsampled to 2 Hz. The cameras, each with a 60° field of view (FOV), offer 360° visual coverage with varied resolutions (front/back: 2880x1860 pixels; sides: 1920x1240 pixels).
To accurately estimate the vehicle's state and position, we also record signals from an inertial measurement unit (IMU) and a real-time kinematic global navigation satellite system (RTK-GNSS) receiver.
During data acquisition, all sensors and control area network (CAN) signals are recorded with synchronized timestamps.
Following the nuScenes~\cite{nuscenes}, the recorded sequences are segmented into 20-second video clips. Further details about the sensor setup are provided in Appendix~\ref{appendix:sensor_setup}.

\begin{figure*}[t]
    \centering
    \includegraphics[width=0.99\linewidth]{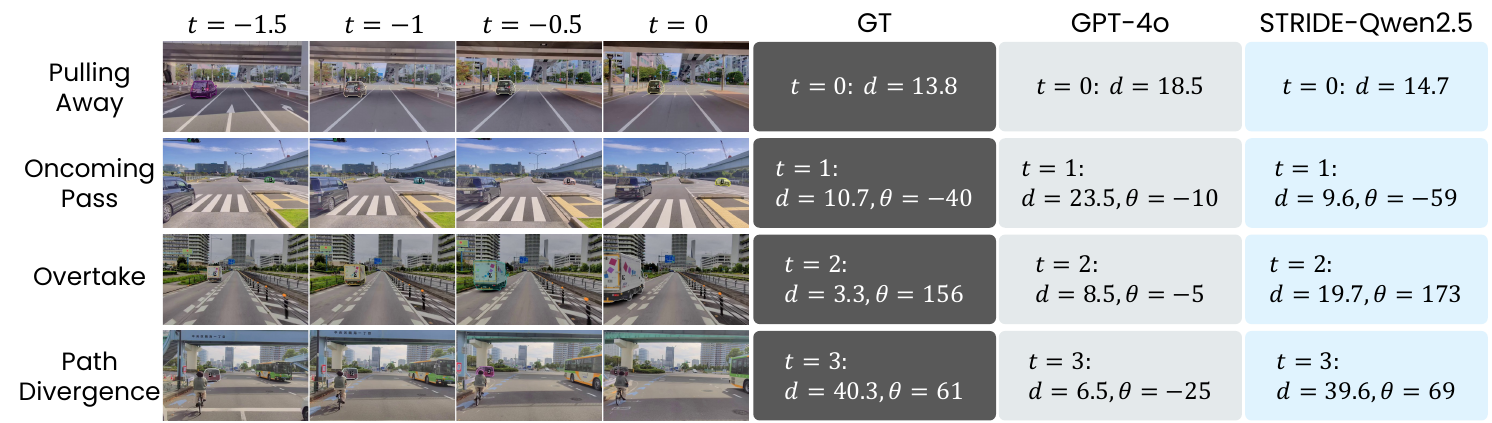}
    \caption{Qualitative results on STRIDE-QA Bench. Across four driving scenarios, our fine-tuned STRIDE-Qwen2.5-VL-7B (labeled as STRIDE-Qwen2.5) consistently delivers more accurate distance and angle estimates than GPT-4o. Note that model responses are abbreviated to highlight key numerical predictions ($t$: time [s], $d$: distance [m], $\theta$: angle [$^\circ$]).}
    \label{fig:qualitative_results_strideqa_bench}
\end{figure*}

\subsection{Annotation Pipeline}
\label{subsec:pipeline}

We propose an automated annotation pipeline that processes sequences of synchronized multi-view RGB images and LiDAR point clouds to generate spatiotemporal question-answer pairs, each consisting of a question, an answer, and supporting visual evidence (e.g., bounding boxes or segmentation masks).

Our pipeline, illustrated in Figure~\ref{fig:overview}, comprises seven components: keyframe sampling, 3D object detection, multi-object tracking, attributes extraction, semantic segmentation, visibility filtering, and question generation.

\noindent\textbf{Keyframe Sampling.}\quad
In our pipeline, a keyframe serves as the temporal anchor for each spatiotemporal question. We sample these at 1 Hz from the 2 to 17-second interval of each clip, which provides each keyframe with a temporal context window spanning from 2 seconds prior to 3 seconds after.

\noindent\textbf{3D Object Detection.}\quad
We adopt BEVFusion~\cite{liu2022bevfusion}, which fuses LiDAR and multi-camera images to perform high-precision 3D object detection. For each keyframe, we estimate the 3D position, orientation, and size of all visible objects. The set of object classes includes those from nuScenes as well as additional custom classes, totaling 45 categories.

\noindent\textbf{Object Tracking.}\quad
To capture temporal dynamics, we track the movement of each object over time. PubTracker~\cite{yin2021center} is selected for its simplicity and efficiency in point-based 3D object tracking, which enables fast and consistent ID assignment across frames without relying on appearance features. We associate 3D bounding boxes across consecutive frames and assign consistent instance IDs to each object. As a result, all frames, including context and future frames associated with each keyframe, are annotated with 3D bounding boxes, and objects can be consistently tracked over time.

\noindent\textbf{Attributes Extractor}\quad
To support spatiotemporal QA generation, we extract relevant attributes for each object in every frame. Based on the 3D bounding boxes and the pose of the ego vehicle, we compute the Euclidean distance and heading angle in the ego-centric coordinate system. In addition to these numerical values, we also generate discrete and qualitative spatial descriptions such as ``left'' or ``1 o'clock'' derived from these quantities.

Moreover, by leveraging the temporally consistent object tracks obtained during the tracking stage, we estimate the velocity of each object from the change in the center position of its bounding box over time. These attributes, including distance, heading, and velocity, are associated with each object in every frame and form the foundation for generating spatiotemporal QA pairs.

\noindent\textbf{Semantic Segmentation.}\quad
To obtain precise 2D segmentation masks for downstream visibility filtering and to provide fine-grained, pixel-level visual grounding for VLMs, we apply SAM 2.1 Large~\cite{ravi2024sam2}. The model generates a class-agnostic mask within each object's projected 3D bounding box. Each mask is then associated with the object's consistent instance ID from the tracking stage.

\noindent\textbf{Visibility Filtering.}\quad
Since bounding boxes and segmentation masks are automatically generated, it is important to filter out unreliable objects to ensure annotation quality. We apply the following three filtering rules: (1) the Intersection-over-Union (IoU) between bounding boxes must be greater than 0.3, (2) the coverage rate must exceed 0.8, and (3) the IoU score predicted by SAM 2.1 Large must also exceed 0.8. By applying these rules, we effectively eliminate noisy detections and retain only high-quality annotations.

\noindent\textbf{Question Generator.}\quad
Finally, we generate template-based QA pairs for each keyframe based on the previously extracted object attributes, such as distance, heading, and velocity. Each object is referred to as \texttt{Region [X]} in the question text, where the number \texttt{X} corresponds to the visual identifier shown in the image. These QA pairs are aligned with the 2D bounding boxes and segmentation masks from the current, context, and future frames. Each QA pair is associated with a keyframe \(t_k\), and refers to a target object or object pair within a temporal window \([t_k - 2, t_k + 3]\). The spatial grounding \(s\) consists of 2D bounding boxes and segmentation masks across the relevant frames \(t\), aligned via consistent instance IDs. As illustrated in Figure~\ref{fig:data_samples}, this results in a dataset composed of paired image, segmentation, and QA examples.

These stages automatically yield spatiotemporal QA pairs from synchronized multi-camera RGB and LiDAR data, ensuring temporally aligned, instance-consistent annotations for VLM training and evaluation; annotation quality details are in Appendix \ref{appendix:dataset}.

\section{Experiments}
\label{sec:experiments}

We evaluate whether STRIDE-QA improves VLMs’ spatiotemporal reasoning in traffic scenes by benchmarking multiple models on spatial and spatiotemporal QA tasks.

\subsection{Experimental Setup}\label{subsec:exp_settings}

\noindent\textbf{Fine-tuned Models.}\quad
We fine-tune two open-source VLMs, Qwen2.5-VL-7B-Instruct and Cosmos-Reason1-7B, on the training split of STRIDE-QA using only the front-camera RGB sequence (\(t \in \{-1.5, -1.0, -0.5, 0\}\,\text{s})\). Side and rear camera images and LiDAR are omitted to match the evaluation setup detailed in the next subsection. We refer to the resulting models as STRIDE-Qwen2.5-VL-7B and STRIDE-Cosmos-Reason1-7B.

\noindent\textbf{Baseline Models.}\quad
We evaluate our fine-tuned models against a range of VLMs. To estimate the upper bound of general-purpose VLMs, we include proprietary GPT-4o and GPT-4o mini~\cite{gpt4technicalreport}. We further compare with the open-source models InternVL2.5-8B~\cite{internvl} and Qwen2.5-VL-7B-Instruct~\cite{qwen2_5}, the spatially enhanced SpatialRGPT-8B~\cite{cheng2024spatialrgpt}, and autonomous driving-specific VLMs, Senna-VLM~\cite{senna} and Cosmos-Reason1-7B~\cite{nvidia2025cosmos-reason1}.
Details of the training process are provided in Appendix~\ref{appendix:implementation_details}.

\noindent\textbf{Spatial Benchmark.}\quad
We evaluate spatial reasoning on the outdoor split of the SpatialRGPT-Bench~\cite{cheng2024spatialrgpt}.
The details of the SpatialRGPT-Bench are provided in Appendix~\ref{appendix:spatialrgpt-bench}.

\noindent\textbf{Spatiotemporal Benchmark.}\quad
Beyond spatial reasoning, we introduce \textbf{STRIDE-QA Bench} to evaluate spatiotemporal reasoning capabilities. We employ three primary types of metrics, which are briefly described in Section~\ref{subsec:spatiotemporal_bench} and detailed in Appendix~\ref{appendix:strideqa-bench}.

\subsection{STRIDE-QA Bench}
\label{subsec:spatiotemporal_bench}

STRIDE-QA Bench is an evaluation suite for spatiotemporal reasoning in urban driving scenes. Beyond spatial QA benchmarks, we assess the performance of spatiotemporal QA tasks defined in Section \ref{sec:preliminary}.
We report four metrics: Localization Success Rate (LSR), Mean Localization Success Rate (MLSR), Temporal Localization Consistency (TLC), and diagnostic per-dimension success rates (SR).

\paragraph{Evaluation Setup.}
The model is given a sequence of four RGB frames from a front-facing onboard camera with a 60° FOV, captured at \(t \in \{-1.5, -1.0, -0.5, 0\}\,\text{s}\). In this sequence, the model observes a single \emph{target agent} from one of six classes (\textit{car}, \textit{large\_vehicle}, \textit{bus}, \textit{pedestrian}, \textit{motorcycle}, or \textit{bicycle}), identified by its segmentation mask in all four frames to provide sufficient context. The model is tasked with predicting the following quantities at \(t \in \{0, 1, 2, 3\}\,\text{s}\): the target agent's \emph{distance}, \emph{velocity}, and \emph{heading angle}, along with the ego vehicle's \emph{velocity}. The agent's heading angle is defined in the ego vehicle's frame of reference, where \(0^\circ\) is forward and positive angles are counter-clockwise on the range \((-180^\circ, 180^\circ]\). This definition is explicitly provided in the prompt for fair evaluation. Notably, the task includes predicting agents that may move out of the camera's FOV at \(t > 0\).

\paragraph{Evaluation Dataset.}

The evaluation dataset is built from held-out recording dates of our STRIDE-QA corpus to ensure train-test separation. We define a \textit{scene group} as a sequence centered on a single target agent observed across four context frames ($t \in {-1.5, -1.0, -0.5, 0}$s) and evaluated across four future timesteps ($t \in {0, 1, 2, 3}$s).
Each scene group includes 13 QA pairs, totaling 5,317 QA pairs across 409 scene groups. To focus on challenging spatiotemporal reasoning, we filter the data to exclude static and repetitive scenes, retaining only those with dynamic interactions. These scene groups are categorized into six dynamic scenarios: \textit{Oncoming Pass}, \textit{Maintain State}, \textit{Overtake}, \textit{Path Divergence}, \textit{Pulling Away From Ego}, and \textit{Minor Relations}. We define an out-of-view (OOV) event where the target object completely exits the front camera's FOV in any future frame ($t \in {1, 2, 3}$s). These scenarios exhibit a clear divide in their OOV likelihood; for instance, scenarios like \textit{Oncoming Pass} and \textit{Overtake} consistently involve OOV events, whereas others like \textit{Maintain State} rarely do. Detailed statistics for all scenarios are provided in Appendix~\ref{appendix:strideqa-bench}.

\begin{table}[t]
  \centering
  \footnotesize
  \setlength{\tabcolsep}{4pt}
  \resizebox{0.45\textwidth}{!}{%
    \begin{tabular}{lcccccc}
      \toprule
      \multirow{2}{*}{Model} & \multicolumn{2}{c}{Original$\,\uparrow$}
      & \multicolumn{2}{c}{Obj. Spatial$\,\uparrow$}
      & \multicolumn{2}{c}{Ego Spatial$\,\uparrow$} \\
      \cmidrule(lr){2-3}\cmidrule(lr){4-5}\cmidrule(lr){6-7}
      & Qual. & Quant. & Qual. & Quant. & Qual. & Quant. \\
      \midrule
      GPT‑4o                                & \textbf{80.5} & 32.5 & \textbf{68.1} & 39.4 & 55.7 & 27.7 \\
      GPT‑4o mini                           & 54.7          & 30.6 & 56.3 & 32.1 & 57.1 & 44.4 \\
      Intern‑VL 2.5 8B                      & 64.1          & 18.8 & 48.4 & 26.6 & 36.3 & 29.1 \\
      \cellcolor{gray!20}Qwen2.5‑VL 7B-Instruct      & 67.2          & 24.4 & \underline{64.0} & 12.8 & 47.1 & 29.3 \\
      \midrule
      SpatialRGPT‑VILA-1.5-8B                        & \underline{75.0} & \textbf{46.9} & 58.4 & 42.2 & 25.3 & 16.9 \\
      \midrule
      Senna‑VLM                           & 18.0 & 0.63 & 9.14 & 4.59 & 5.88 & 2.03 \\
      \cellcolor{gray!20}Cosmos‑Reason1‑7B  & 53.9          & 30.0 & 55.2 & 21.1 & 33.2 & 20.3 \\
      \midrule
      \cellcolor{cyan!10} STRIDE-Qwen2.5-VL-7B     & 69.5 & \underline{37.5} & 61.1 & \textbf{61.5} & \underline{77.9} & \textbf{70.3} \\
      \cellcolor{cyan!10} STRIDE-Cosmos-Reason1-7B & 71.1 & 30.0 & 62.2 & \underline{58.7} & \textbf{79.9} & \underline{68.9} \\
      \bottomrule
    \end{tabular}
  }
  \caption{SpatialRGPT-Bench results. Evaluation of qualitative (Qual.) and quantitative (Quant.) spatial reasoning performance on the original SpatialRGPT dataset, as well as our object-centric and ego-centric extensions.}
  \label{tab:spatialrgpt_bench}
\end{table}

\paragraph{Localization Success Rate (LSR).}
LSR is a binary metric that evaluates whether a model's spatial prediction is simultaneously accurate in both distance and direction.
Here, $\hat d_t$ and $\hat\theta_t$ denote model predictions, while $d_t^{*}$ and $\theta_t^{*}$ are the corresponding ground-truth values.
A prediction is successful only if the estimated distance and orientation, expressed in polar coordinates $\hat{\mathbf{p}}_t=(\hat d_t,\hat\theta_t)$, satisfy
\[
|\hat d_t-d_t^{\!*}| < 0.25\,d_t^{\!*}
\quad\text{and}\quad
|\hat\theta_t-\theta_t^{\!*}| < 10^{\circ}.
\]
We adopt a $\pm25\%$ distance margin following prior work\,\cite{cheng2024spatialrgpt} and set the $\pm10^{\circ}$ heading margin so that at 10\,m the lateral deviation equals a standard 3.5\,m lane width.

\paragraph{Mean Localization Success Rate (MLSR).}
MLSR averages the per-frame LSR over a sequence, providing a softer measure of temporal stability:
\[
\text{MLSR} = \frac{1}{|G|} \sum_{g\in G} \frac{1}{T+1} \sum_{t=0}^{T} s_{g,t},
\]
where $\mathbf{s}_{g}=[s_{g,0},\dots,s_{g,T}]$ denotes the LSR success bits for scene $g\in G$.

\paragraph{Temporal Localization Consistency (TLC).}
TLC is a strict metric that measures the proportion of sequences where the agent is successfully localized across all four timesteps, i.e., the target is never lost:
\[
\text{TLC} = \frac{1}{|G|} \sum_{g\in G} \mathbf{1}\!\bigl[s_{g,0} \land s_{g,1} \land s_{g,2} \land s_{g,3}\bigr].
\]
where $\mathbf{s}_{g}=[s_{g,0},\dots,s_{g,T}]$ is the sequence of LSR success bits for a scene group $g$ with $T=3$.

\paragraph{Per-dimension Success Rates.}
Single-axis success rates for distance, heading angle, and velocities are computed for diagnostic analysis. Full metric definitions and detailed results are provided in Appendix \ref{appendix:strideqa-bench}.

\subsection{Results on SpatialRGPT-Bench}

Table~\ref{tab:spatialrgpt_bench} summarizes the results. STRIDE-Qwen2.5-VL-7B and STRIDE-Cosmos-Reason1-7B substantially outperform their base models across all splits. In the quantitative Object-centric Spatial QA, STRIDE-Qwen2.5-VL-7B rises from $12.8$ to $61.5$ ($+48.7~\text{pt}$, $\times 4.8$) and STRIDE-Cosmos-Reason1-7B from $21.1$ to $58.7$ ($+37.6~\text{pt}$, $\times 2.8$). Similar gains appear in the quantitative Ego-centric split, where scores improve from $29.3$ to $70.3$ ($+41.0~\text{pt}$) and from $20.3$ to $68.9$ ($+48.6~\text{pt}$) for the two models, respectively. These results indicate that training on the STRIDE-QA dataset boosts fine-grained spatial reasoning on external benchmarks. Although GPT-4o and SpatialRGPT-8B achieve the highest accuracy on the original split, their performance drops markedly on the Object-centric subset. This gap suggests that generic large models do not automatically transfer to camera-centric setups, possibly due to viewpoint shifts and annotation noise. A detailed ablation is provided in Appendix~\ref{appendix:spatialrgpt-bench}.

\begin{table}[t]
  \centering
  \footnotesize
  \setlength{\tabcolsep}{4pt}
  \resizebox{0.45\textwidth}{!}{%
  \begin{tabular}{lcccccc}
    \toprule
    \multirow{2}{*}{Model} &
    \multicolumn{4}{c}{$\text{LSR}\uparrow$} &
    \multirow{2}{*}{$\text{MLSR}\uparrow$} &
    \multirow{2}{*}{$\text{TLC}\uparrow$} \\
    \cmidrule(lr){2-5}
     & 0s & 1s & 2s & 3s \\
    \midrule
    GPT-4o         & 18.1 & 6.6 & 6.1 & 7.6 & 9.6 & 0.7 \\
    GPT-4o mini    & 4.6 & 2.0 & 0.7 & 0.7 & 2.0 & 0.0 \\
    InternVL2.5-8B & 2.4 & 1.0 & 1.7 & 0.7 & 1.5 & 0.0 \\
    \cellcolor{gray!20}Qwen2.5-VL-7B-Instruct & 1.0 & 3.4 & 4.4 & 1.0 & 2.4 & 0.0 \\
    \midrule
    SpatialRGPT-VILA-1.5-8B & 0.5 & 0.2 & 0.2 & 0.0 & 0.2 & 0.0 \\
    \midrule
    Senna-VLM & 1.0 & 0.0 & 0.2 & 0.0 & 0.3 & 0.0 \\
    \cellcolor{gray!20}Cosmos-Reason1-7B & 1.5 & 3.2 & 2.0 & 1.5 & 2.0 & 0.0 \\
    \midrule
    \cellcolor{cyan!10}STRIDE-Qwen2.5-VL-7B & \underline{96.3} & \textbf{46.2} & \textbf{38.4} & \textbf{38.9} & \textbf{55.0} & \textbf{28.4} \\
    \cellcolor{cyan!10}STRIDE-Cosmos-Reason1-7B & \textbf{96.8} & \underline{43.5} & \underline{37.4} & \underline{36.2} & \underline{53.5} & \underline{25.4} \\
    \bottomrule
  \end{tabular}
  }
  \caption{STRIDE-QA Bench results. our fine-tuned models achieve the best performance across LSR@t, MSLR, and TLC metrics.}
  \label{tab:stride_bench_lsr}
\end{table}

\subsection{Results on STRIDE-QA Bench}

As presented in Table~\ref{tab:stride_bench_lsr}, all baseline models, including powerful general-purpose VLMs, exhibit poor performance on the spatiotemporal benchmark defined in Section~\ref{subsec:spatiotemporal_bench}. Their Localization Success Rate (LSR) scores are low, and they achieve near-zero results on the multi-frame (MLSR) and temporal consistency (TLC) metrics. This indicates a fundamental inability of existing models to perform reliable spatiotemporal reasoning in complex driving scenarios.

In contrast, models fine-tuned on STRIDE-QA demonstrate dramatic performance gains. Our top-performing model,
STRIDE-Qwen2.5-VL-7B achieves an $LSR_{t=0}$ of 96.3\% (96$\times$ baseline), demonstrating precise spatial understanding. This extends to future horizons with an $LSR_{t=3}$ of 38.9\% (39$\times$ baseline).
Furthermore, it obtains an MLSR of $55.0$ and a TLC of $28.4$, confirming that our dataset effectively teaches consistent reasoning across viewpoints and time (see Figure~\ref{fig:qualitative_results_strideqa_bench} for qualitative examples).

However, the results also highlight the task’s difficulty. While a TLC score of $28.4$ improves from $0$, it indicates that achieving short-term consistent reasoning remains challenging. This suggests that fine-tuning on our dataset helps, but may be insufficient to capture the complexity of real-world dynamic prediction.

In conclusion, our work makes two key contributions. First, we show that fine-tuning on STRIDE-QA closes a critical gap in spatiotemporal reasoning capabilities of VLMs. Second, by quantifying remaining challenges in short-term consistency, our benchmark guides future research toward building robust, safety-critical autonomous systems.

\begin{figure}[t]
    \centering
    \includegraphics[width=0.99\linewidth]{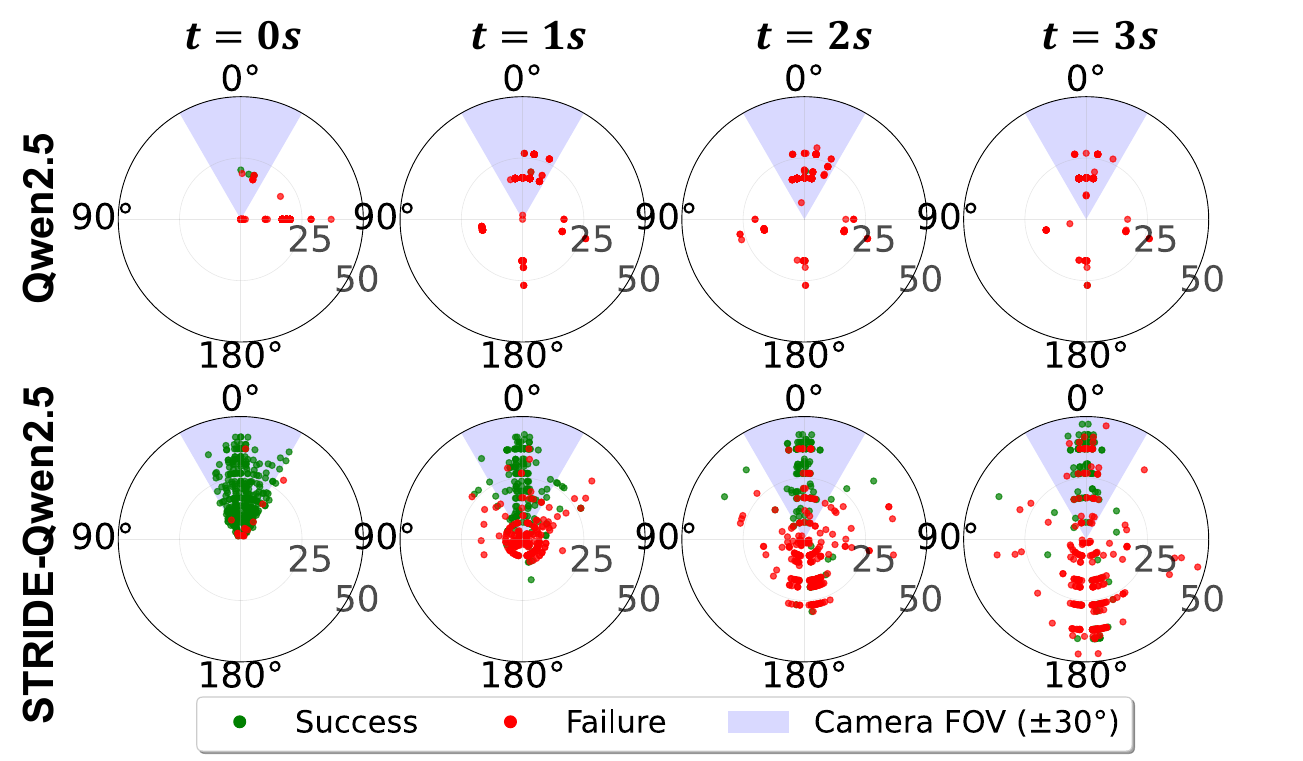}
    \caption{
    Comparison of localization (angle=heading, radius=ego distance).
    Finetuning (bottom) boosts performance vs. base (top): dense, plausible forecasts and near-perfect $t=0\,s$ localization.
    Green/red dots indicate LSR success/failure; the blue wedge is the camera $\pm30^{\circ}$ FOV.
    }
    \label{fig:rsr_polar_plot}
\end{figure}

\section{Analysis}
\label{sec:analysis}

Beyond performance metrics, we conduct in-depth analysis to better understand the strengths, limitations, and generalization behaviors of VLMs trained with STRIDE-QA.

\subsection{Qualitative Analysis of Prediction Patterns}
A qualitative analysis of prediction patterns in Figure~\ref{fig:rsr_polar_plot} reveals fundamental differences in model behavior. The baseline VLM (top row) exhibits a consistent failure mode: its predictions are not only sparse but also systematically biased, repeating similar incorrect guesses regardless of the visual input. This suggests the model defaults to a simplistic, memorized behavior instead of grounding its reasoning in the visual context. In contrast, our fine-tuned model (bottom row) is responsive, generating dense predictions across the 360-degree space and accurately localizing objects in the present frame ($t=0s$). The baseline's failure to produce context-aware, continuous predictions demonstrates a lack of temporal consistency, which is a fundamental prerequisite for motion planning. This prerequisite is often overlooked in prior work connecting VLMs to planning modules~\cite{DriveLM, tian2024drivevlm}, a gap our framework is designed to address by providing a framework to explicitly train and evaluate this skill.

\subsection{Error Analysis on Out-of-View Prediction}
\label{subsec:oov}
\begin{figure}[t]
    \centering
    \includegraphics[width=\linewidth]{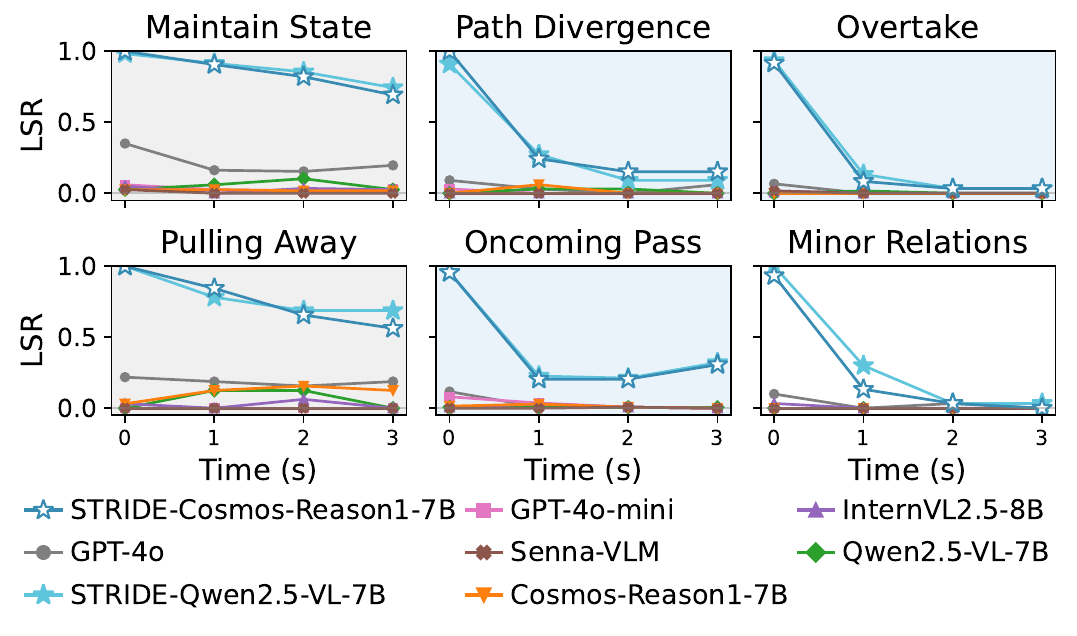}
    \caption{LSR trends across dynamic scenarios, grouped by out-of-view (OOV) likelihood. The sharp performance drop in scenarios with a high OOV likelihood (blue background) compared to in-view scenarios (grey background) highlights that OOV prediction is the primary challenge.
    }
    \label{fig:lsr_per_scenario}
\end{figure}

To understand the root cause of the temporal degradation identified in the previous section, we analyze our model's performance across the different dynamic scenarios defined in our benchmark (Figure~\ref{fig:lsr_per_scenario}). A clear pattern emerges: in scenarios where the target agent is likely to remain within the camera's FOV, such as \textit{Maintain State}, the LSR declines gracefully. Conversely, in scenarios where the agent tends to exit the FOV, such as \textit{Overtake}, \textit{Oncoming Pass}, and \textit{Path Divergence}, the LSR degrades much more sharply.
This stark contrast strongly suggests that the primary failure mode for long-term prediction is the model's inability to reason about OOV object trajectories. This finding exposes a key limitation of relying on single-camera input for forecasting and highlights that integrating multi-camera information is a critical direction to build robust autonomous driving.

\section{Conclusion}
\label{sec:conclusion}

We introduced STRIDE-QA, a large-scale VQA dataset addressing spatiotemporal reasoning gaps in autonomous-driving VLMs. Fine-tuning on it boosted performance: our best model reached 55.0\% MLSR and 28.4 TLC, dramatically surpassing otherwise near-zero baselines. STRIDE-QA establishes a foundation for robust, physically grounded vision-language understanding in autonomous systems.

\section*{Acknowledgments}

This paper is based on results obtained from GENIAC (Generative AI Accelerator Challenge, a project to strengthen Japan's generative AI development capabilities), a project (JPNP20017) implemented by the Ministry of Economy, Trade and Industry (METI) and the New Energy and Industrial Technology Development Organization (NEDO).

\bibliography{aaai2026}

@String(CVPR= {IEEE Conf. Comput. Vis. Pattern Recog.})

@String(ECCV= {Eur. Conf. Comput. Vis.})

@String(ICLR = {Int. Conf. Learn. Represent.})

@String(AAAI = {AAAI})

@String(CVPR  = {CVPR})

@String(ECCV  = {ECCV})

@String(ICLR  = {ICLR})

@misc{senna,
    title={Senna: Bridging Large Vision-Language Models and End-to-End Autonomous Driving}, 
    author={Bo Jiang and Shaoyu Chen and Bencheng Liao and Xingyu Zhang and Wei Yin and Qian Zhang and Chang Huang and Wenyu Liu and Xinggang Wang},
    year={2024},
    eprint={2410.22313},
    archivePrefix={arXiv},
    primaryClass={cs.CV},
    url={https://arxiv.org/abs/2410.22313}, 
}

@inproceedings{tian2024drivevlm,
    title={DriveVLM: The Convergence of Autonomous Driving and Large Vision-Language Models},
    author={Xiaoyu Tian and Junru Gu and Bailin Li and Yicheng Liu and Yang Wang and Zhiyong Zhao and Kun Zhan and Peng Jia and XianPeng Lang and Hang Zhao},
    booktitle={8th Annual Conference on Robot Learning (CoRL)},
    year={2024},
    url={https://openreview.net/forum?id=928V4Umlys}
}

@inproceedings{DriveLM,
    title = {DriveLM: Driving with Graph Visual Question Answering},
    author = {Sima, Chonghao and Renz, Katrin and Chitta, Kashyap and Chen, Li and Zhang, Hanxue and Xie, Chengen and Bei\ss{}wenger, Jens and Luo, Ping and Geiger, Andreas and Li, Hongyang},
    booktitle = {Proceedings of the European Conference on Computer Vision (ECCV)},
    year = {2024},
    month = {September},
    pages = {256–274}
}

@inproceedings{lingoqa,
    title = {LingoQA: Visual Question Answering for Autonomous Driving},
    author = {Marcu, Ana-Maria and Chen, Long and H\"{u}nermann, Jan and Karnsund, Alice and Hanotte, Benoit and Chidananda, Prajwal and Nair, Saurabh and Badrinarayanan, Vijay and Kendall, Alex and Shotton, Jamie and Arani, Elahe and Sinavski, Oleg},
    booktitle = {Proceedings of the European Conference on Computer Vision (ECCV)},
    pages = {252--269},
    year = {2024}
}

@inproceedings{nuscenes,
  title={nuScenes: A Multimodal Dataset for Autonomous Driving},
  author={Caesar, Holger and Bankiti, Varun and Lang, Alex H and Vora, Sourabh and Liong, Venice Erin and Xu, Qiang and Krishnan, Anush and Pan, Yu and Baldan, Giancarlo and Beijbom, Oscar},
  booktitle={Proceedings of the IEEE/CVF Conference on Computer Vision and Pattern Recognition (CVPR)},
  pages={11621--11631},
  year={2020}
}

@inproceedings{cheng2024spatialrgpt,
  title={SpatialRGPT: Grounded Spatial Reasoning in Vision-Language Models},
  author={Cheng, An-Chieh and Yin, Hongxu and Fu, Yang and Guo, Qiushan and Yang, Ruihan and Kautz, Jan and Wang, Xiaolong and Liu, Sifei},
  booktitle={NeurIPS},
  year={2024}
}

@inproceedings{chen2024spatialvlm,
  title={Spatialvlm: Endowing vision-language models with spatial reasoning capabilities},
  author={Chen, Boyuan and Xu, Zhuo and Kirmani, Sean and Ichter, Brain and Sadigh, Dorsa and Guibas, Leonidas and Xia, Fei},
  booktitle={Proceedings of the IEEE/CVF Conference on Computer Vision and Pattern Recognition (CVPR)},
  pages={14455--14465},
  year={2024}
}

@misc{rt2,
      title={RT-2: Vision-Language-Action Models Transfer Web Knowledge to Robotic Control}, 
      author={Anthony Brohan and Noah Brown and Justice Carbajal and Yevgen Chebotar and Xi Chen and Krzysztof Choromanski and Tianli Ding and Danny Driess and Avinava Dubey and Chelsea Finn and Pete Florence and Chuyuan Fu and Montse Gonzalez Arenas and Keerthana Gopalakrishnan and Kehang Han and Karol Hausman and Alexander Herzog and Jasmine Hsu and Brian Ichter and Alex Irpan and Nikhil Joshi and Ryan Julian and Dmitry Kalashnikov and Yuheng Kuang and Isabel Leal and Lisa Lee and Tsang-Wei Edward Lee and Sergey Levine and Yao Lu and Henryk Michalewski and Igor Mordatch and Karl Pertsch and Kanishka Rao and Krista Reymann and Michael Ryoo and Grecia Salazar and Pannag Sanketi and Pierre Sermanet and Jaspiar Singh and Anikait Singh and Radu Soricut and Huong Tran and Vincent Vanhoucke and Quan Vuong and Ayzaan Wahid and Stefan Welker and Paul Wohlhart and Jialin Wu and Fei Xia and Ted Xiao and Peng Xu and Sichun Xu and Tianhe Yu and Brianna Zitkovich},
      year={2023},
      eprint={2307.15818},
      archivePrefix={arXiv},
      primaryClass={cs.RO},
      url={https://arxiv.org/abs/2307.15818}, 
}

@misc{black2024pi0visionlanguageactionflowmodel,
      title={$\pi_0$: A Vision-Language-Action Flow Model for General Robot Control}, 
      author={Kevin Black and Noah Brown and Danny Driess and Adnan Esmail and Michael Equi and Chelsea Finn and Niccolo Fusai and Lachy Groom and Karol Hausman and Brian Ichter and Szymon Jakubczak and Tim Jones and Liyiming Ke and Sergey Levine and Adrian Li-Bell and Mohith Mothukuri and Suraj Nair and Karl Pertsch and Lucy Xiaoyang Shi and James Tanner and Quan Vuong and Anna Walling and Haohuan Wang and Ury Zhilinsky},
      year={2024},
      eprint={2410.24164},
      archivePrefix={arXiv},
      primaryClass={cs.LG},
      url={https://arxiv.org/abs/2410.24164}, 
}

@inproceedings{tod3cap,
    author = {Jin, Bu and Zheng, Yupeng and Li, Pengfei and Li, Weize and Zheng, Yuhang and Hu, Sujie and Liu, Xinyu and Zhu, Jinwei and Yan, Zhijie and Sun, Haiyang and Zhan, Kun and Jia, Peng and Long, Xiaoxiao and Chen, Yilun and Zhao, Hao},
    title = {TOD3Cap: Towards 3D Dense Captioning in Outdoor Scenes},
    year = {2024},
    isbn = {978-3-031-72648-4},
    publisher = {Springer-Verlag},
    address = {Berlin, Heidelberg},
    url = {https://doi.org/10.1007/978-3-031-72649-1_21},
    doi = {10.1007/978-3-031-72649-1_21},
    booktitle = {Computer Vision – ECCV 2024: 18th European Conference, Milan, Italy, September 29 – October 4, 2024, Proceedings, Part XVIII},
    pages = {367–384},
    numpages = {18},
    location = {Milan, Italy}
}

@InProceedings{blink,
    author="Fu, Xingyu
    and Hu, Yushi
    and Li, Bangzheng
    and Feng, Yu
    and Wang, Haoyu
    and Lin, Xudong
    and Roth, Dan
    and Smith, Noah A.
    and Ma, Wei-Chiu
    and Krishna, Ranjay",
    editor="Leonardis, Ale{\v{s}}
    and Ricci, Elisa
    and Roth, Stefan
    and Russakovsky, Olga
    and Sattler, Torsten
    and Varol, G{\"u}l",
    title="BLINK: Multimodal Large Language Models Can See but Not Perceive",
    booktitle="Computer Vision -- ECCV 2024",
    year="2025",
    publisher="Springer Nature Switzerland",
    address="Cham",
    pages="148--166",
    isbn="978-3-031-73337-6"
}

@misc{gpt4technicalreport,
      title={GPT-4 Technical Report}, 
      author={OpenAI and Josh Achiam and Steven Adler and Sandhini Agarwal and Lama Ahmad and Ilge Akkaya and Florencia Leoni Aleman and Diogo Almeida and Janko Altenschmidt and Sam Altman and Shyamal Anadkat and Red Avila and Igor Babuschkin and Suchir Balaji and Valerie Balcom and Paul Baltescu and Haiming Bao and Mohammad Bavarian and Jeff Belgum and Irwan Bello and Jake Berdine and Gabriel Bernadett-Shapiro and Christopher Berner and Lenny Bogdonoff and Oleg Boiko and Madelaine Boyd and Anna-Luisa Brakman and Greg Brockman and Tim Brooks and Miles Brundage and Kevin Button and Trevor Cai and Rosie Campbell and Andrew Cann and Brittany Carey and Chelsea Carlson and Rory Carmichael and Brooke Chan and Che Chang and Fotis Chantzis and Derek Chen and Sully Chen and Ruby Chen and Jason Chen and Mark Chen and Ben Chess and Chester Cho and Casey Chu and Hyung Won Chung and Dave Cummings and Jeremiah Currier and Yunxing Dai and Cory Decareaux and Thomas Degry and Noah Deutsch and Damien Deville and Arka Dhar and David Dohan and Steve Dowling and Sheila Dunning and Adrien Ecoffet and Atty Eleti and Tyna Eloundou and David Farhi and Liam Fedus and Niko Felix and Simón Posada Fishman and Juston Forte and Isabella Fulford and Leo Gao and Elie Georges and Christian Gibson and Vik Goel and Tarun Gogineni and Gabriel Goh and Rapha Gontijo-Lopes and Jonathan Gordon and Morgan Grafstein and Scott Gray and Ryan Greene and Joshua Gross and Shixiang Shane Gu and Yufei Guo and Chris Hallacy and Jesse Han and Jeff Harris and Yuchen He and Mike Heaton and Johannes Heidecke and Chris Hesse and Alan Hickey and Wade Hickey and Peter Hoeschele and Brandon Houghton and Kenny Hsu and Shengli Hu and Xin Hu and Joost Huizinga and Shantanu Jain and Shawn Jain and Joanne Jang and Angela Jiang and Roger Jiang and Haozhun Jin and Denny Jin and Shino Jomoto and Billie Jonn and Heewoo Jun and Tomer Kaftan and Łukasz Kaiser and Ali Kamali and Ingmar Kanitscheider and Nitish Shirish Keskar and Tabarak Khan and Logan Kilpatrick and Jong Wook Kim and Christina Kim and Yongjik Kim and Jan Hendrik Kirchner and Jamie Kiros and Matt Knight and Daniel Kokotajlo and Łukasz Kondraciuk and Andrew Kondrich and Aris Konstantinidis and Kyle Kosic and Gretchen Krueger and Vishal Kuo and Michael Lampe and Ikai Lan and Teddy Lee and Jan Leike and Jade Leung and Daniel Levy and Chak Ming Li and Rachel Lim and Molly Lin and Stephanie Lin and Mateusz Litwin and Theresa Lopez and Ryan Lowe and Patricia Lue and Anna Makanju and Kim Malfacini and Sam Manning and Todor Markov and Yaniv Markovski and Bianca Martin and Katie Mayer and Andrew Mayne and Bob McGrew and Scott Mayer McKinney and Christine McLeavey and Paul McMillan and Jake McNeil and David Medina and Aalok Mehta and Jacob Menick and Luke Metz and Andrey Mishchenko and Pamela Mishkin and Vinnie Monaco and Evan Morikawa and Daniel Mossing and Tong Mu and Mira Murati and Oleg Murk and David Mély and Ashvin Nair and Reiichiro Nakano and Rajeev Nayak and Arvind Neelakantan and Richard Ngo and Hyeonwoo Noh and Long Ouyang and Cullen O'Keefe and Jakub Pachocki and Alex Paino and Joe Palermo and Ashley Pantuliano and Giambattista Parascandolo and Joel Parish and Emy Parparita and Alex Passos and Mikhail Pavlov and Andrew Peng and Adam Perelman and Filipe de Avila Belbute Peres and Michael Petrov and Henrique Ponde de Oliveira Pinto and Michael and Pokorny and Michelle Pokrass and Vitchyr H. Pong and Tolly Powell and Alethea Power and Boris Power and Elizabeth Proehl and Raul Puri and Alec Radford and Jack Rae and Aditya Ramesh and Cameron Raymond and Francis Real and Kendra Rimbach and Carl Ross and Bob Rotsted and Henri Roussez and Nick Ryder and Mario Saltarelli and Ted Sanders and Shibani Santurkar and Girish Sastry and Heather Schmidt and David Schnurr and John Schulman and Daniel Selsam and Kyla Sheppard and Toki Sherbakov and Jessica Shieh and Sarah Shoker and Pranav Shyam and Szymon Sidor and Eric Sigler and Maddie Simens and Jordan Sitkin and Katarina Slama and Ian Sohl and Benjamin Sokolowsky and Yang Song and Natalie Staudacher and Felipe Petroski Such and Natalie Summers and Ilya Sutskever and Jie Tang and Nikolas Tezak and Madeleine B. Thompson and Phil Tillet and Amin Tootoonchian and Elizabeth Tseng and Preston Tuggle and Nick Turley and Jerry Tworek and Juan Felipe Cerón Uribe and Andrea Vallone and Arun Vijayvergiya and Chelsea Voss and Carroll Wainwright and Justin Jay Wang and Alvin Wang and Ben Wang and Jonathan Ward and Jason Wei and CJ Weinmann and Akila Welihinda and Peter Welinder and Jiayi Weng and Lilian Weng and Matt Wiethoff and Dave Willner and Clemens Winter and Samuel Wolrich and Hannah Wong and Lauren Workman and Sherwin Wu and Jeff Wu and Michael Wu and Kai Xiao and Tao Xu and Sarah Yoo and Kevin Yu and Qiming Yuan and Wojciech Zaremba and Rowan Zellers and Chong Zhang and Marvin Zhang and Shengjia Zhao and Tianhao Zheng and Juntang Zhuang and William Zhuk and Barret Zoph},
      year={2024},
      eprint={2303.08774},
      archivePrefix={arXiv},
      primaryClass={cs.CL},
      url={https://arxiv.org/abs/2303.08774}, 
}

@INPROCEEDINGS{KITTI,
  author={Geiger, Andreas and Lenz, Philip and Urtasun, Raquel},
  booktitle={2012 IEEE Conference on Computer Vision and Pattern Recognition}, 
  title={Are we ready for autonomous driving? The KITTI vision benchmark suite}, 
  year={2012},
  volume={},
  number={},
  pages={3354-3361},
  doi={10.1109/CVPR.2012.6248074}}

@misc{nvidia2025cosmos-reason1,
      title={Cosmos-Reason1: From Physical Common Sense To Embodied Reasoning}, 
      author={NVIDIA and : and Alisson Azzolini and Hannah Brandon and Prithvijit Chattopadhyay and Huayu Chen and Jinju Chu and Yin Cui and Jenna Diamond and Yifan Ding and Francesco Ferroni and Rama Govindaraju and Jinwei Gu and Siddharth Gururani and Imad El Hanafi and Zekun Hao and Jacob Huffman and Jingyi Jin and Brendan Johnson and Rizwan Khan and George Kurian and Elena Lantz and Nayeon Lee and Zhaoshuo Li and Xuan Li and Tsung-Yi Lin and Yen-Chen Lin and Ming-Yu Liu and Alice Luo and Andrew Mathau and Yun Ni and Lindsey Pavao and Wei Ping and David W. Romero and Misha Smelyanskiy and Shuran Song and Lyne Tchapmi and Andrew Z. Wang and Boxin Wang and Haoxiang Wang and Fangyin Wei and Jiashu Xu and Yao Xu and Xiaodong Yang and Zhuolin Yang and Xiaohui Zeng and Zhe Zhang},
      year={2025},
      eprint={2503.15558},
      archivePrefix={arXiv},
      primaryClass={cs.AI},
      url={https://arxiv.org/abs/2503.15558}, 
}

@misc{dashcam_anonymizer,
  author={Gupta, Varun},
  year={2023},
  title={Dashcam Anonymizer},
  howpublished = {\url{https://github.com/varungupta31/dashcam_anonymizer}},
  note = {Accessed: 2025-08-31}
}

@inproceedings{liu2022bevfusion,
  title={BEVFusion: Multi-Task Multi-Sensor Fusion with Unified Bird's-Eye View Representation},
  author={Liu, Zhijian and Tang, Haotian and Amini, Alexander and Yang, Xingyu and Mao, Huizi and Rus, Daniela and Han, Song},
  booktitle={IEEE International Conference on Robotics and Automation (ICRA)},
  year={2023}
}

@InProceedings{yin2021center,
    author    = {Yin, Tianwei and Zhou, Xingyi and Krahenbuhl, Philipp},
    title     = {Center-Based 3D Object Detection and Tracking},
    booktitle = {Proceedings of the IEEE/CVF Conference on Computer Vision and Pattern Recognition (CVPR)},
    month     = {June},
    year      = {2021},
    pages     = {11784-11793}
}

@inproceedings{ravi2024sam2,
    title={SAM 2: Segment Anything in Images and Videos},
    author={Nikhila Ravi and Valentin Gabeur and Yuan-Ting Hu and Ronghang Hu and Chaitanya Ryali and Tengyu Ma and Haitham Khedr and Roman R{\"a}dle and Chloe Rolland and Laura Gustafson and Eric Mintun and Junting Pan and Kalyan Vasudev Alwala and Nicolas Carion and Chao-Yuan Wu and Ross Girshick and Piotr Dollar and Christoph Feichtenhofer},
    booktitle={The Thirteenth International Conference on Learning Representations (ICLR)},
    year={2025},
    url={https://openreview.net/forum?id=Ha6RTeWMd0}
}

@inproceedings{qian2024nuscenesqa,
  title={Nuscenes-qa: A multi-modal visual question answering benchmark for autonomous driving scenario},
  author={Qian, Tianwen and Chen, Jingjing and Zhuo, Linhai and Jiao, Yang and Jiang, Yu-Gang},
  booktitle={Proceedings of the AAAI Conference on Artificial Intelligence},
  volume={38},
  pages={4542--4550},
  year={2024}
}

@article{hu2022lora,
  title={Lora: Low-rank adaptation of large language models.},
  author={Hu, Edward J and Shen, Yelong and Wallis, Phillip and Allen-Zhu, Zeyuan and Li, Yuanzhi and Wang, Shean and Wang, Lu and Chen, Weizhu and others},
  journal={ICLR},
  volume={1},
  number={2},
  pages={3},
  year={2022}
}

@misc{qwen2_5,
      title={Qwen2.5-VL Technical Report}, 
      author={Shuai Bai and Keqin Chen and Xuejing Liu and Jialin Wang and Wenbin Ge and Sibo Song and Kai Dang and Peng Wang and Shijie Wang and Jun Tang and Humen Zhong and Yuanzhi Zhu and Mingkun Yang and Zhaohai Li and Jianqiang Wan and Pengfei Wang and Wei Ding and Zheren Fu and Yiheng Xu and Jiabo Ye and Xi Zhang and Tianbao Xie and Zesen Cheng and Hang Zhang and Zhibo Yang and Haiyang Xu and Junyang Lin},
      year={2025},
      eprint={2502.13923},
      archivePrefix={arXiv},
      primaryClass={cs.CV},
      url={https://arxiv.org/abs/2502.13923}, 
}

@inproceedings{llava,
author      = {Liu, Haotian and Li, Chunyuan and Wu, Qingyang and Lee, Yong Jae},
title       = {Visual Instruction Tuning},
booktitle   = {NeurIPS},
year        = {2023}
}

@inproceedings{internvl,
    title={Internvl: Scaling up vision foundation models and aligning for generic visual-linguistic tasks},
    author={Chen, Zhe and Wu, Jiannan and Wang, Wenhai and Su, Weijie and Chen, Guo and Xing, Sen and Zhong, Muyan and Zhang, Qinglong and Zhu, Xizhou and Lu, Lewei and others},
    booktitle={Proceedings of the IEEE/CVF Conference on Computer Vision and Pattern Recognition (CVPR)},
    pages={24185--24198},
    year={2024}
}

@inproceedings{hypersim,
  title={Hypersim: A photorealistic synthetic dataset for holistic indoor scene understanding},
  author={Roberts, Mike and Ramapuram, Jason and Ranjan, Anurag and Kumar, Atulit and Bautista, Miguel Angel and Paczan, Nathan and Webb, Russ and Susskind, Joshua M},
  booktitle={Proceedings of the IEEE/CVF international conference on computer vision},
  pages={10912--10922},
  year={2021}
}

@misc{baruch2021arkitscenes,
  title         = {Arkitscenes: A Diverse Real-World Dataset for 3D Indoor Scene Understanding Using Mobile RGB-D Data},
  author        = {Baruch, Gilad and Chen, Zhuoyuan and Dehghan, Afshin and Dimry, Tal and Feigin, Yuri and Fu, Peter and Gebauer, Thomas and Joffe, Brandon and Kurz, Daniel and Schwartz, Arik and others},
  year          = {2021},
  eprint        = {2111.08897},
  archivePrefix = {arXiv},
  howpublished  = {\url{https://arxiv.org/abs/2111.08897}}
}

@misc{som,
  title         = {Set-of-mark Prompting Unleashes Extraordinary Visual Grounding in GPT-4V},
  author        = {Yang, Jianwei and Zhang, Hao and Li, Feng and Zou, Xueyan and Li, Chunyuan and Gao, Jianfeng},
  year          = {2023},
  eprint        = {2310.11441},
  archivePrefix = {arXiv},
  howpublished  = {\url{https://arxiv.org/abs/2310.11441}},
}

@inproceedings{zhou2025tumtraf,
    title={{TUMT}raf Video{QA}: Dataset and Benchmark for Unified Spatio-Temporal Video Understanding in Traffic Scenes},
    author={Xingcheng Zhou and Konstantinos Larintzakis and Hao Guo and Walter Zimmer and Mingyu Liu and Hu Cao and Jiajie Zhang and Venkatnarayanan Lakshminarasimhan and Leah Strand and Alois Knoll},
    booktitle={Forty-second International Conference on Machine Learning},
    year={2025},
    url={https://openreview.net/forum?id=Yfoi5O68rf}
}

@article{nuprompt, 
    title={Language Prompt for Autonomous Driving}, volume={39}, url={https://ojs.aaai.org/index.php/AAAI/article/view/32902},
    doi={10.1609/aaai.v39i8.32902},
    number={8},
    journal={Proceedings of the AAAI Conference on Artificial Intelligence},
    author={Wu, Dongming and Han, Wencheng and Liu, Yingfei and Wang, Tiancai and Xu, Cheng-Zhong and Zhang, Xiangyu and Shen, Jianbing}, 
    year={2025},
    pages   = {8359--8367},
}

@misc{nuplanqa,
      title={NuPlanQA: A Large-Scale Dataset and Benchmark for Multi-View Driving Scene Understanding in Multi-Modal Large Language Models}, 
      author={Sung-Yeon Park and Can Cui and Yunsheng Ma and Ahmadreza Moradipari and Rohit Gupta and Kyungtae Han and Ziran Wang},
      year={2025},
      eprint={2503.12772},
      archivePrefix={arXiv},
      primaryClass={cs.CV},
      url={https://arxiv.org/abs/2503.12772}, 
}

@inproceedings{goyal2017making,
  title={Making the v in vqa matter: Elevating the role of image understanding in visual question answering},
  author={Goyal, Yash and Khot, Tejas and Summers-Stay, Douglas and Batra, Dhruv and Parikh, Devi},
  booktitle={Proceedings of the IEEE conference on computer vision and pattern recognition},
  pages={6904--6913},
  year={2017}
}

@misc{ainslie2023gqa,
  title          = {GQA: Training Generalized Multi-Query Transformer Models from Multi-Head Checkpoints},
  author         = {Ainslie, Joshua and Lee-Thorp, James and De~Jong, Michiel and Zemlyanskiy, Yury and Lebr{\'o}n, Federico and Sanghai, Sumit},
  year           = {2023},
  eprint         = {2305.13245},
  archivePrefix  = {arXiv},
}

@inproceedings{wu2023referring,
    title={Referring Multi-Object Tracking},
    author= {Wu, Dongming and Han, Wencheng and Wang, Tiancai and Dong, Xingping and Zhang, Xiangyu and Shen, Jianbing},
    booktitle={Proceedings of the IEEE/CVF Conference on Computer Vision and Pattern Recognition (CVPR)},
    pages={14633--14642},
    year={2023},
}

@inproceedings{Wu_2023_CVPR,
    author    = {Wu, Dongming and Han, Wencheng and Wang, Tiancai and Dong, Xingping and Zhang, Xiangyu and Shen, Jianbing},
    title     = {Referring Multi-Object Tracking},
    booktitle = {Proceedings of the IEEE/CVF Conference on Computer Vision and Pattern Recognition (CVPR)},
    month     = {June},
    year      = {2023},
    pages     = {14633-14642}
}

@inproceedings{sunrggbd,
    author = { Song, Shuran and Lichtenberg, Samuel P. and Xiao, Jianxiong },
    booktitle = { 2015 IEEE Conference on Computer Vision and Pattern Recognition (CVPR) },
    title = { SUN RGB-D: A RGB-D scene understanding benchmark suite },
    year = {2015},
    volume = {},
    ISSN = {1063-6919},
    pages = {567-576},
    doi = {10.1109/CVPR.2015.7298655},
    url = {https://doi.ieeecomputersociety.org/10.1109/CVPR.2015.7298655},
    publisher = {IEEE Computer Society},
    address = {Los Alamitos, CA, USA},
}

\clearpage

\appendix
\label{appendix}

\section{STRIDE-QA Dataset Details}
\label{appendix:dataset}

\subsection{Sensor Setup and Calibration}
\label{appendix:sensor_setup}

\begin{figure}[ht]
    \centering
    \includegraphics[width=0.99\linewidth]{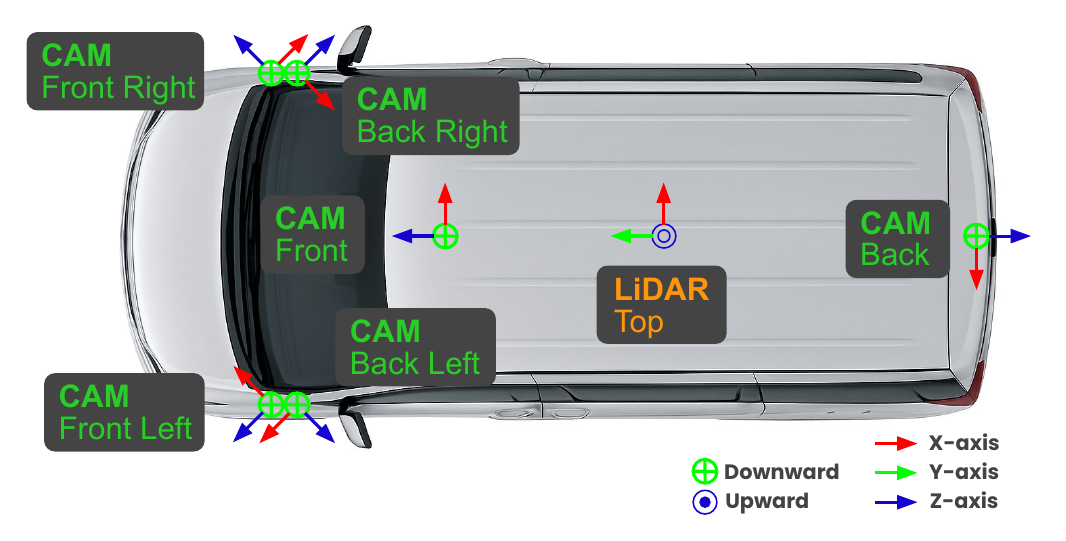}
    \caption{Sensor setup of our data collection platform.}
    \label{fig:alphard}
\end{figure}

\noindent\textbf{Sensor Setup}\quad
The six cameras and LiDAR sensor are mounted as illustrated in Figure \ref{fig:alphard}. This camera configuration provides 360-degree visual coverage.

\noindent\textbf{Sensor Calibration}\quad
To achieve a high-quality multi-sensor dataset, careful calibration of sensor intrinsic and extrinsic parameters is required. We perform a comprehensive calibration process involving LiDAR, cameras, and VRTK (Vision-RTK). The key steps are described below:

\begin{itemize}
    \item \textbf{High-Precision Mapping and Trajectory Estimation} \\
    We employ LiDAR SLAM (Simultaneous Localization and Mapping) to estimate both the 3D point cloud map and the vehicle's trajectory. The initial pose is provided by the VRTK system, which integrates RTK-GNSS, IMU, and camera data. The generated map is then aligned to a global reference map using Iterative Closest Point (ICP).

    \item \textbf{Relative Pose Estimation: VRTK, LiDAR, and Ground} \\
    We estimate the VRTK-LiDAR pose by aligning their trajectories via rigid transform fitting, fixing translation and optimizing rotation. The LiDAR-ground pose is determined by detecting the ground plane from the point cloud and computing the median height, pitch, and roll across frames.

    \item \textbf{Camera Calibration and Scale Alignment} \\
    Structure from Motion (SfM) is used to recover camera intrinsics and inter-camera extrinsics. To resolve scale ambiguity, we align the SfM reconstruction with the metric LiDAR map using a 7-DoF scale-aware ICP, initialized from trajectory comparison.
\end{itemize}

This procedure yields the final extrinsic parameters between all sensors and the intrinsic parameters for each camera.

\subsection{Statistics}
We provide statistics of the STRIDE-QA dataset to highlight its scale and content diversity.
Table~\ref{tab:dataset_qa_category_stats} shows the distribution of qualitative and quantitative QA pairs.
In addition, Table~\ref{tab:sampled_class_distribution} presents the distribution of object classes across QA categories, showing that many questions target common traffic participants such as vehicles, motorcycles, and pedestrians. Object counts in the table are computed as unique objects within each conversation thread rather than per QA pair. Consequently, the same physical object may appear in multiple threads and be counted more than once across the dataset.

\begin{table}[ht]
    \centering
    \footnotesize
    \resizebox{0.45\textwidth}{!}{
    \begin{tabular}{l rr rr r}
        \toprule
        \multirow{2}{*}{QA Category} & \multicolumn{2}{c}{Qualitative} & \multicolumn{2}{c}{Quantitative} & \multirow{2}{*}{Total} \\
        \cmidrule(lr){2-3} \cmidrule(lr){4-5}
         & train & val & train & val & \\
        \midrule
        Object-centric Spatial QA
          & 2.20 & 0.11
          & 1.10 & 0.06
          & 3.47 \\
        Ego-centric Spatial QA
          & 3.10 & 0.16
          & 4.33 & 0.23
          & 7.82 \\
        Ego-centric Spatiotemporal QA
          & -- & --
          & 4.73 & 0.40
          & 5.13 \\
        \midrule
        \textbf{Total}
          & 5.30 & 0.28
          & 10.17 & 0.69
          & 16.43 \\
        \bottomrule
    \end{tabular}
    }
    \caption{Distribution of QA pairs across categories and splits. Values are shown in millions (M).}
    \label{tab:dataset_qa_category_stats}
\end{table}

\begin{table}[ht]
  \centering
  \begin{tabular}{l r r r}
    \toprule
    \textbf{Class} & \textbf{Obj. (S)} & \textbf{Ego. (S)} & \textbf{Ego. (ST)} \\
    \midrule
    vehicle        & 424,196 & 440,278 & 261,453 \\
    pedestrian     & 136,738 & 139,205 & 38,990 \\
    large vehicle  & 110,755   & 119,585 & 65,118 \\
    bicycle        & 38,047   & 40,830   & 12,462 \\
    bus            & 20,817   & 23,427   & 14,769 \\
    motorcycle     & 17,120   & 18,276   & 8,501 \\
    kick scooter   & 453      & 473     & 169 \\
    \bottomrule
  \end{tabular}
  \caption{Distribution of object classes across QA categories in the STRIDE-QA dataset. Obj. (S): Object-centric Spatial QA, Ego. (S): Ego-centric Spatial QA, Ego. (ST): Ego-centric Spatiotemporal QA. Note that a single object instance can be associated with questions in multiple categories; therefore, the columns are not mutually exclusive.}
  \label{tab:sampled_class_distribution}
\end{table}

\subsection{Pipeline Validation and Quality Assessment}
\label{subsec:annotation_quality_assessment}

Given the vast scale of the STRIDE-QA dataset (16 million QA pairs), a direct manual evaluation of all annotations is infeasible. We therefore adopt an indirect yet rigorous approach: evaluating the performance of the core components that determine the pipeline's quality, namely 3D object detection and multi-object tracking.

\begin{figure*}[ht]
    \centering
    \includegraphics[width=0.99\linewidth]{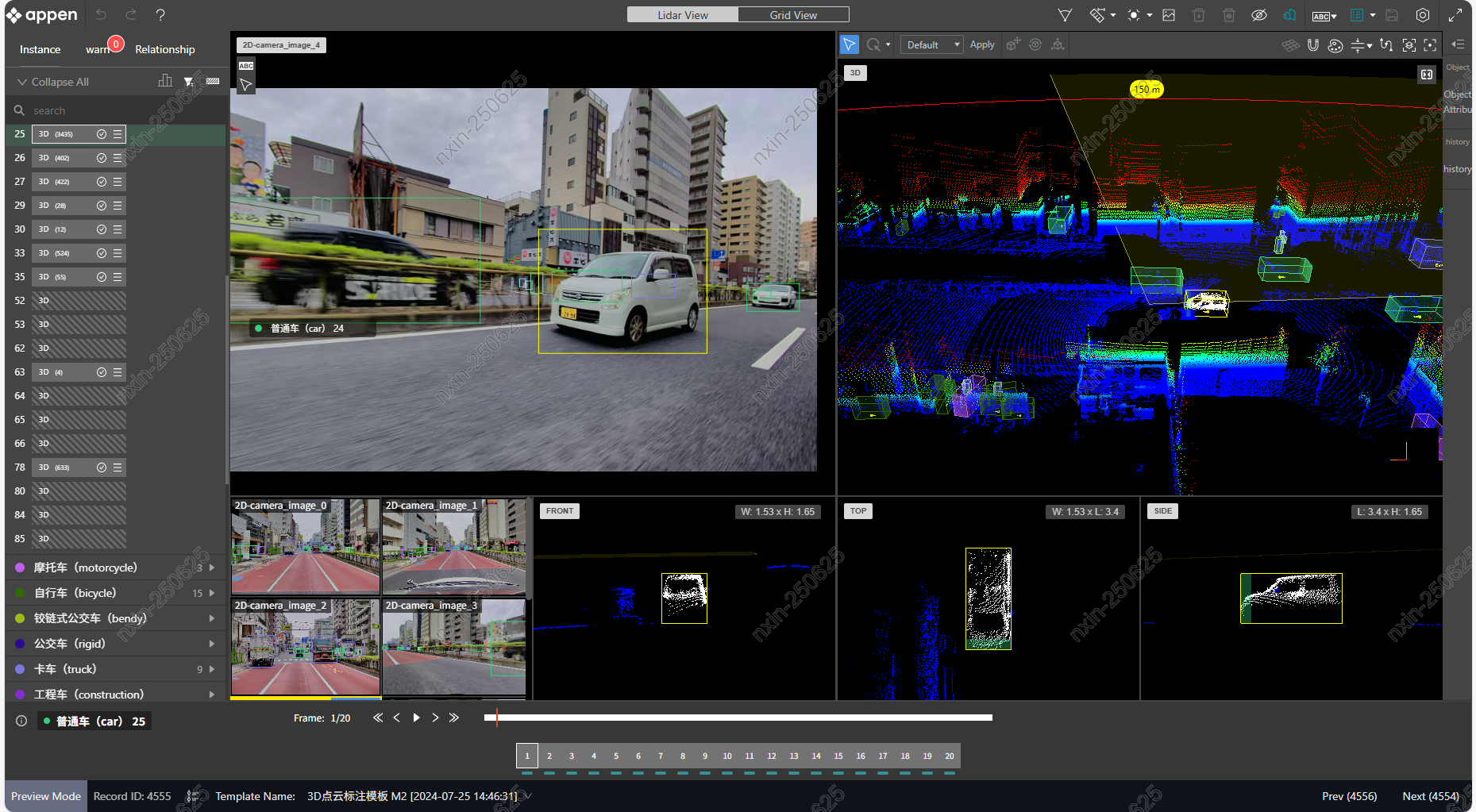}
    \caption{Illustration of representative 2D and 3D annotations produced with Appen® MatrixGo on our driving dataset, showing synchronized camera imagery and LiDAR point clouds with class-specific bounding boxes.}
    \label{fig:appen_annotation_tool_example_1}
\end{figure*}

For this evaluation, we first prepared a ground truth dataset collected separately from the main STRIDE-QA driving logs but with an identical sensor configuration. The creation of this dataset was outsourced to expert human annotators using the Appen® MatrixGo tool (shown in Figure~\ref{fig:appen_annotation_tool_example_1}) and was carried out based on our detailed guidelines. All submitted annotations were reviewed and corrected by our team to ensure quality. The annotations follow the nuScenes format, with each 3D bounding box assigned a consistent instance ID tracked throughout each 10-second scene. This dataset is split into a training set of 1688 scenes (approximately 281 minutes) and a validation set of 100 scenes (approx. 17 minutes).

Using this 100-scene validation set as ground truth, we evaluate detection performance with five standard metrics (mAP, mATE, mASE, mAOE, mAVE) and tracking performance with four metrics (AMOTA, AMOTP, Recall, IDSW). The results are presented in Table~\ref{tab:quality-assessment}.

\begin{table}[ht]
  \centering
  \setlength{\tabcolsep}{6pt}
  \begin{tabular}{l l r l}
    \toprule
    \textbf{Task}      & \textbf{Metric}            & \textbf{Score} & \textbf{Range} \\
    \midrule
    \multirow{5}{*}{Detection} %
                       & mAP\,$\uparrow$            & 0.701 & $[0, 1]$ \\
                       & mATE\,$\downarrow$ (m)     & 0.136 & $[0, \infty)$ \\
                       & mASE\,$\downarrow$         & 0.168 & $[0, 1]$ \\
                       & mAOE\,$\downarrow$ (rad)   & 0.146 & $[0, \pi]$ \\
                       & mAVE\,$\downarrow$ (m/s)   & 1.280 & $[0, \infty)$ \\
    \midrule
    \multirow{4}{*}{Tracking}  %
                       & AMOTA\,$\uparrow$          & 0.676 & $[0, 1]$ \\
                       & AMOTP\,$\downarrow$ (m)    & 0.556 & $[0, \infty)$ \\
                       & Recall\,$\uparrow$         & 0.700 & $[0, 1]$ \\
                       & IDSW\,$\downarrow$         & 687   & $[0, \infty)$ \\
    \bottomrule
  \end{tabular}
  \caption{Class-averaged detection and tracking metrics achieved by our automated annotation pipeline.}
  \label{tab:quality-assessment}
\end{table}

\noindent\textbf{Detection Quality.}\quad
The detector achieves a high precision with an mAP (IoU@0.5:0.95) of 0.701 and a translational error (mATE) of only 0.136\,m (13.6 cm). Shape (mASE=0.168) and heading (mAOE=0.146 rad $\approx$ 8.4$^{\circ}$) are likewise accurate, and the velocity error is kept sufficiently low (mAVE=1.28 m/s). These numbers indicate that the values embedded in STRIDE-QA questions are derived from centimeter-level geometric estimates. The $\pm$25\% distance tolerance is based on prior work~\cite{cheng2024spatialrgpt}, while the $\pm$10$^{\circ}$ angular tolerance is a practical threshold set so that the lateral error at a 10m distance corresponds to a standard lane width (approx. 3.5m).

\noindent\textbf{Tracking Quality.}\quad
The overall multi-object tracking (MOT) performance achieves a solid score of AMOTA = 0.676 (evaluated on 100 scenes containing 10,461 unique object tracks). This metric provides a comprehensive evaluation of tracking failures (ID switches), missed detections (Recall), and false positives, and this score indicates our pipeline maintains a good balance between these errors.
For our task of generating QA pairs from stable tracks, we emphasize two metrics: tracking reliability and positional accuracy. To this end, our system achieves high reliability with only 687 ID switches out of 10,461 total tracks (approx. 6.9 per 10-second scene). For positional accuracy, the Average Multi-Object Tracking Precision (AMOTP) is kept to 0.556\,m. This high positional accuracy is crucial for generating spatially precise questions and answers, as their content relies on accurate object locations.

\noindent\textbf{Summary.}\quad
In conclusion, the proposed pipeline achieves high positional accuracy, with an average detection error (mATE) of under 14\,cm and an average tracking error (AMOTP) of under 0.6\,m. The results of this validation provide strong backing for the claim that the STRIDE-QA dataset combines large-scale QA pairs with the geometric precision necessary for the rigorous evaluation of spatiotemporal reasoning in VLMs.

\subsection{Privacy and Anonymization}
We anonymize the entire dataset by automatically blurring faces and vehicle license plates using Dashcam Anonymizer~\cite{dashcam_anonymizer}, and by removing all identifiable timestamp metadata while preserving the original frame order.

\section{Implementation Details}
\label{appendix:implementation_details}

We fine-tune two open-source VLMs, Qwen2.5-VL-7B~\cite{qwen2_5} and Cosmos-Reason1-7B~\cite{nvidia2025cosmos-reason1}, on the training split of the STRIDE-QA dataset. Each training sample consists of four context frames, resized to \(532 \times 336\) pixels, with Region ID-annotated masks generated by the Set-of-Mark (SoM) method~\cite{som}. We employ the LoRA~\cite{hu2022lora} method for parameter-efficient fine-tuning. All models are trained on 16 NVIDIA H100 GPUs using DeepSpeed (ZeRO Stage 2) for distributed training optimization, under the same hyperparameter settings summarized in Table~\ref{tab:hyperparameters}. All training runs use a fixed random seed of 42 to ensure reproducibility. During evaluation, input frames are pre-processed in the same manner as during training. We use a decoding temperature of 0 to ensure deterministic inference.

\begin{table}[h!]
\centering
\begin{tabular}{ll}
    \toprule
    \textbf{Parameter} & \textbf{Value} \\
    \midrule
    \multicolumn{2}{l}{\textit{Training Configuration}} \\
    \quad Epochs & 2 \\
    \quad Global Batch Size & 64 \\
    \quad Precision & bfloat16 \\
    \midrule
    \multicolumn{2}{l}{\textit{Optimizer (AdamW)}} \\
    \quad Learning Rate (LLM) & 5e-5 \\
    \quad Learning Rate (Vision Encoder) & 2e-6 \\
    \quad Learning Rate (Projection Head) & 1e-5 \\
    \quad Betas ($\beta_1, \beta_2$) & (0.9, 0.999) \\
    \quad Epsilon ($\epsilon$) & 1e-8 \\
    \quad Weight Decay & 0.1 \\
    \midrule
    \multicolumn{2}{l}{\textit{Scheduler}} \\
    \quad Type & Cosine \\
    \quad Warmup Ratio & 0.05 \\
    \midrule
    \multicolumn{2}{l}{\textit{LoRA Configuration}} \\
    \quad Rank & 16 \\
    \quad Alpha & 32 \\
    \quad Dropout & 0.05 \\
    \quad Bias & none \\
    \bottomrule
\end{tabular}
\caption{Hyperparameters for fine-tuning.}
\label{tab:hyperparameters}
\end{table}

\section{The STRIDE-QA Bench}
\label{appendix:strideqa-bench}

This section provides further details on the STRIDE-QA Bench.

\subsection{Evaluation Setup and Prompt}

The evaluation task involves taking four consecutive context frames (\(t \in \{-1.5, -1.0, -0.5, 0\}\,\text{s}\)) as input to predict the target agent's distance, heading angle, and velocity at future timesteps (\(t \in \{0, 1, 2, 3\}\,\text{s}\)). The input includes masks with specified regions generated by the Set-of-Mark (SoM) method~\cite{som}. However, one of the baseline models, SpatialRGPT-VILA-1.5-8B, includes a built-in region extractor and does not support multi-frame inputs. Therefore, to ensure a fair comparison, we applied the same SoM-based approach as used for the other models.

\begin{figure*}[!ht]
    \centering
    \includegraphics[width=0.99\linewidth]{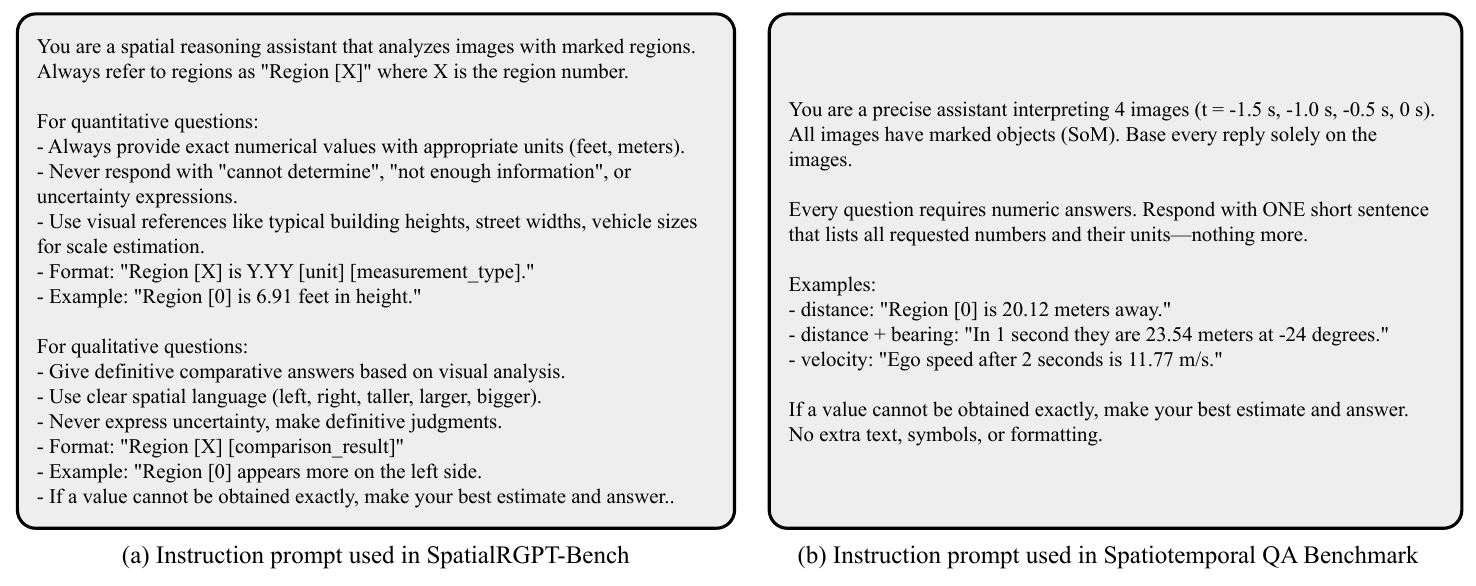}
    \caption{The instruction prompts defining the evaluation task and output format for (a) the SpatialRGPT-Bench and (b) our STRIDE-QA Bench.
    }
    \label{fig:instruction_prompt}
\end{figure*}

The full instruction prompt used for all models during evaluation is provided in Figure~\ref{fig:instruction_prompt} (b).

\subsection{Evaluation Dataset Statistics and Definitions}

\begin{figure*}[ht]
    \centering
    \includegraphics[width=0.95\linewidth]{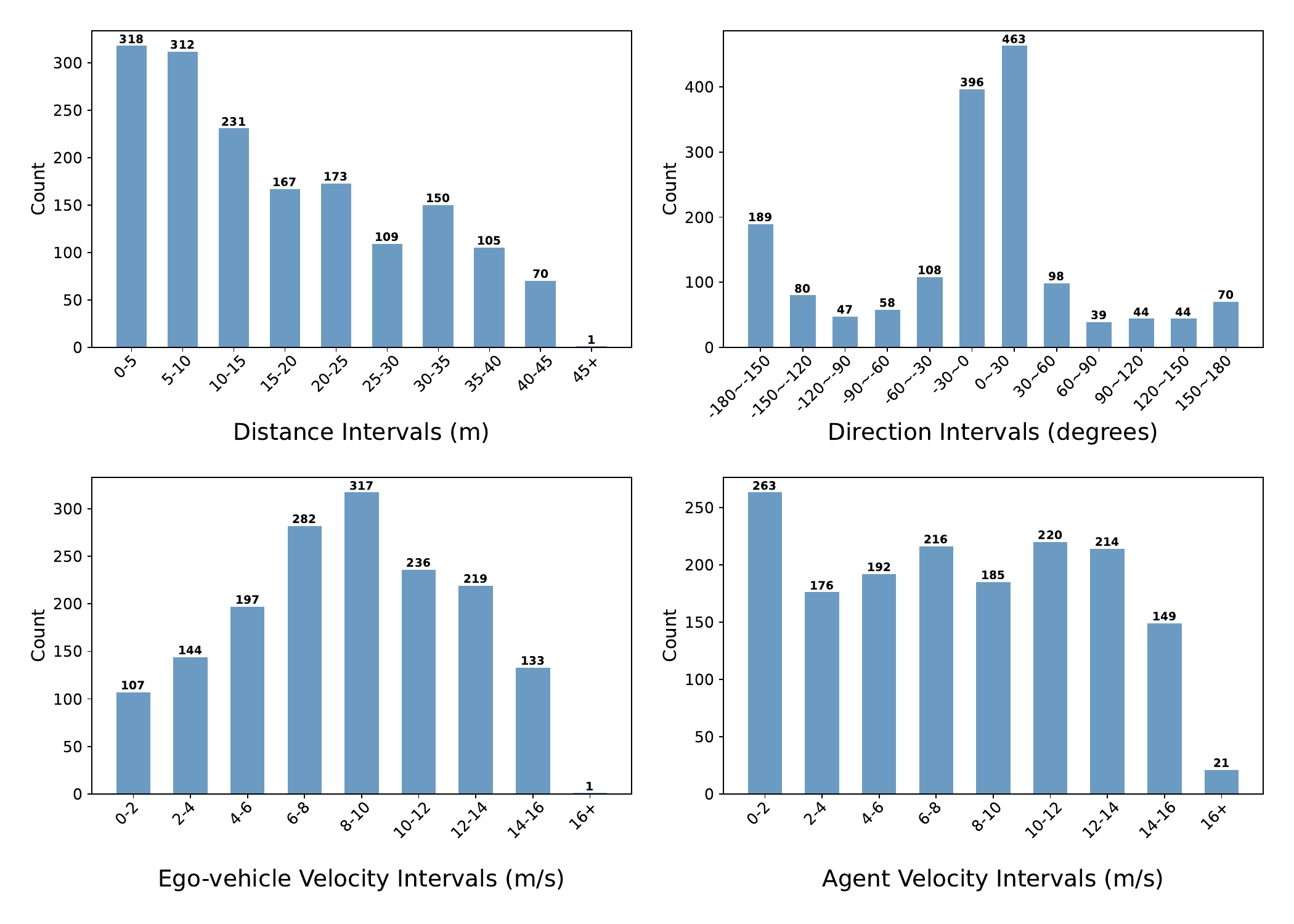}
    \caption{STRIDE-QA Bench sample distribution.}
    \label{fig:st_qa_bench_data_distribution}
\end{figure*}

Our evaluation dataset is composed of 409 unique scene groups. The distribution of these scenes into six dynamic scenarios and their respective Out-of-View (OOV) rates are detailed in Table~\ref{tab:scenario_distribution}. Furthermore, Figure~\ref{fig:st_qa_bench_data_distribution} illustrates the distribution of key physical quantities across the entire benchmark, demonstrating that our dataset covers a diverse range of distances, heading angles, and velocities for both the ego vehicle and target agents. A details of the object classes included in the evaluation is provided in Table~\ref{tab:sampled_class_distribution_st_bench}.

\begin{table}[ht]
  \centering
  \footnotesize
  \setlength{\tabcolsep}{4pt}
  \resizebox{0.45\textwidth}{!}{%
    \begin{tabular}{lrrr}
      \toprule
      Scenario & Total Scenes & Scenes with OOV & OOV Rate \\
      \midrule
      \rowcolor{gray!20} Maintain State          & 117 &   6 & 0.05 \\
      \rowcolor{gray!20} Pulling Away From Ego   &  32 &   6 & 0.19 \\
      \midrule
      \rowcolor{cyan!10} Oncoming Pass           & 137 & 137 & 1.00 \\
      \rowcolor{cyan!10} Overtake                &  60 &  59 & 0.98 \\
      \rowcolor{cyan!10} Path Divergence         &  33 &  31 & 0.94 \\
      \midrule
      Minor Relations         &  30 &  27 & 0.90 \\
      \midrule
      Total                   & 409 & 266 & 0.65 \\
      \bottomrule
    \end{tabular}
  }
  \caption{Distribution of the 409 evaluation scenes by dynamic scenario and out-of-view (OOV) likelihood. The data shows a clear split between in-view (gray background) and OOV-prone scenarios (blue background), with 65\% of all scenes involving OOV. The \textit{Minor} category aggregates scenarios with 30 or fewer samples.}
  \label{tab:scenario_distribution}
\end{table}

\begin{table}[ht]
  \centering
  \begin{tabular}{l r}
    \toprule
    \textbf{Class} & \textbf{Count} \\
    \midrule
    vehicle        & 303 \\
    large vehicle  & 61  \\
    pedestrian     & 17  \\
    bicycle        & 13  \\
    bus            & 8   \\
    motorcycle     & 7   \\
    \midrule
    Total          & 409 \\
    \bottomrule
  \end{tabular}
  \caption{Distribution of object classes in the STRIDE-QA Bench.}
  \label{tab:sampled_class_distribution_st_bench}
\end{table}

Each dynamic scenario is described below:

\noindent\textbf{Oncoming Pass:} The target agent approaches from the opposite direction and passes the ego vehicle. This scenario frequently involves OOV events.

\noindent\textbf{Maintain State:} The target agent and the ego vehicle travel in roughly the same direction and at a similar speed, maintaining a relatively constant spatial relationship.

\noindent\textbf{Overtake:} The ego vehicle overtakes the target agent within the 3-second future horizon. This also frequently involves OOV events.

\noindent\textbf{Path Divergence:} The trajectories of the target agent and the ego vehicle diverge at an intersection, such as when one vehicle turns while the other continues straight. This frequently involves OOV events.

\noindent\textbf{Pulling Away From Ego:} The target agent, traveling ahead of the ego vehicle, accelerates and increases the separation distance.

\noindent\textbf{Minor Relations:} A category that aggregates less frequent or ambiguous scenarios that cannot be cleanly categorized above, such as incomplete overtakes, passing pedestrians, or distant crossing.

\subsection{Metric Definitions}

While the main text reports on integrated metrics such as LSR, MLSR, and TLC, this section provides a supplementary definition for the Per-dimension Success Rate (SR) used for diagnostic analysis. SR evaluates whether the error of a predicted physical quantity falls within a pre-defined tolerance. The tolerance for each physical quantity is set as follows:
\begin{itemize}
    \item \textbf{Distance:} The prediction error must be within $\pm$25\% of the ground truth value.
    \item \textbf{Heading Angle:} The prediction error must be within $\pm$10$^{\circ}$ of the ground truth value.
    \item \textbf{Velocity:} A hybrid tolerance is used; the prediction error must be within $\pm$20\% of the ground truth value, switching to an absolute tolerance of $\pm$0.5 m/s for ground truth speeds below 1.0 m/s.
\end{itemize}

\subsection{Per-dimension Evaluation Results}

Table~\ref{tab:full_sr_results} reports the Per-dimension Success Rate (SR) of each model at prediction horizons.

\begin{table*}[ht]
  \centering
  \scriptsize
  \setlength{\tabcolsep}{4pt}
  \resizebox{0.90\textwidth}{!}{%
  \begin{tabular}{l *{16}{c}}
    \toprule
    \multirow{2}{*}{Model} &
    \multicolumn{4}{c}{$\text{SR}_{d}\uparrow$} &
    \multicolumn{4}{c}{$\text{SR}_{\theta}\uparrow$} &
    \multicolumn{4}{c}{$\text{SR}^{\mathrm{ego}}_{v}\uparrow$} &
    \multicolumn{4}{c}{$\text{SR}^{\mathrm{agent}}_{v}\uparrow$} \\
    \cmidrule(lr){2-5}\cmidrule(lr){6-9}\cmidrule(lr){10-13}\cmidrule(lr){14-17}
     & 0s & 1s & 2s & 3s & 0s & 1s & 2s & 3s & 0s & 1s & 2s & 3s & 0s & 1s & 2s & 3s \\
    \midrule
    GPT-4o & 34.7 & 23.0 & 24.9 & 20.5 & 41.3 & 21.3 & 21.5 & 25.9 & 18.6 & 25.7 & 31.5 & 26.2 & 13.0 & 12.7 & 11.2 & 14.2 \\
    GPT-4o mini & 17.4 & 15.6 & 21.0 & 20.8 & 35.7 & 17.4 & 10.0 & 10.5 & 27.6 & 29.6 & 27.6 & 29.8 & 20.5 & 24.4 & 26.7 & 24.2 \\
    InternVL2.5-8B & 15.9 & 18.6 & 20.8 & 22.0 & 13.9 & 7.8 & 6.1 & 3.2 & 0.5 & 3.7 & 25.9 & 15.2 & 3.4 & 2.7 & 17.8 & 7.6 \\
    \cellcolor{gray!20}Qwen2.5-VL-7B-Instruct & 21.8 & 12.0 & 21.0 & 12.5 & 1.0 & 21.5 & 17.8 & 20.0 & 8.6 & 23.5 & 33.3 & 10.8 & 10.3 & 12.2 & 21.5 & 10.5 \\
    \midrule
    SpatialRGPT-VILA-1.5-8B & 7.3 & 4.2 & 6.8 & 3.2 & 24.0 & 11.0 & 9.0 & 6.1 & 0.0 & 0.2 & 9.3 & 2.0 & 2.7 & 1.2 & 8.3 & 3.4 \\
    \midrule
    Senna-VLM & 11.2 & 2.7 & 6.4 & 4.2 & 3.7 & 0.7 & 1.0 & 0.0 & 4.6 & 5.1 & 7.8 & 2.7 & 0.7 & 3.4 & 6.4 & 4.9 \\
    \cellcolor{gray!20}Cosmos-Reason1-7B & 13.0 & 11.2 & 17.8 & 13.4 & 5.6 & 12.7 & 13.2 & 12.2 & 28.9 & 31.1 & 33.5 & 28.1 & 16.4 & 26.9 & 31.5 & 31.8 \\
    \midrule
    \cellcolor{cyan!10}STRIDE-Qwen2.5-VL-7B & \underline{96.3} & \underline{66.0} & \textbf{51.3} & \textbf{48.9} & \textbf{100} & \textbf{57.9} & \textbf{56.2} & \textbf{61.1} & \textbf{68.0} & \textbf{65.3} & \textbf{63.6} & \textbf{58.9} & \textbf{59.4} & \textbf{60.4} & \underline{53.3} & \textbf{52.8} \\
    \cellcolor{cyan!10}STRIDE-Cosmos-Reason1-7B & \textbf{96.8} & \textbf{69.9} & \underline{49.6} & \underline{47.2} & \textbf{100} & \underline{53.8} & \underline{55.7} & \textbf{61.1} & \underline{62.8} & \underline{59.7} & \underline{58.4} & \underline{54.8} & \textbf{59.4} & \underline{58.2} & \textbf{53.5} & \underline{46.7} \\
    \bottomrule
  \end{tabular}
  }
  \caption{Detailed Per-dimension Success Rate (SR) results on the STRIDE-QA Bench. All values are success rates (\%). Our fine-tuned models (blue background), particularly STRIDE-Qwen2.5-VL-7B, demonstrate a substantial performance advantage over all baselines across every metric and time horizon, especially when compared to their original base models (gray background).
  }
  \label{tab:full_sr_results}
\end{table*}

\section{The SpatialRGPT-Bench}
\label{appendix:spatialrgpt-bench}
The SpatialRGPT-Bench comprises data from outdoor scenes (nuScenes~\cite{nuscenes}, KITTI~\cite{KITTI}), indoor scenes (SUNRGBD~\cite{sunrggbd}, ARKitScenes~\cite{baruch2021arkitscenes}), and simulated scenes (Hypersim~\cite{hypersim}). In this study, we use only the outdoor split to evaluate the spatial understanding of VLMs in urban environments. We exclude QA categories that are less relevant to traffic scenes, such as Below/Above, and omit evaluation items already covered by the STRIDE-QA Bench.
The specific instruction prompt used for this benchmark is shown in Figure~\ref{fig:instruction_prompt}~(a).

Table~\ref{tab:spatialrgpt_details} reports the results for each QA category on SpatialRGPT-Bench. There is a substantial discrepancy between the average scores for the SpatialRGPT-Bench outdoor split and our Object-centric QA split. Because the evaluated QA categories do not align perfectly, a direct comparison is difficult. Nevertheless, two main factors appear to drive the pronounced gap between the SpatialRGPT-Bench outdoor split and the Object-centric QA split. The first factor is annotation error. Our dataset is annotated with a LiDAR- and multi-view sensor-fusion pipeline (see Section \ref{subsec:annotation_quality_assessment}), achieving high geometric accuracy, whereas the original SpatialRGPT-Bench split estimates depth from a single front-view camera, which may produce larger errors. The second factor concerns camera field of view. The Object-centric split uses a narrower FOV than the outdoor split, removing much of the surrounding spatial context and making relative-position reasoning more challenging. These observations further underscore the difficulty of making fair cross-benchmark comparisons when evaluating spatial-understanding capabilities.

\begin{table*}[h]
  \centering
  \renewcommand{\arraystretch}{0.80}
  \setlength{\tabcolsep}{2pt}
  \scriptsize
  \begin{tabular*}{\linewidth}{@{\extracolsep{\fill}} lccccccc}
    \multicolumn{8}{l}{\textbf{Original (outdoor)}} \\
    \toprule
    Model & Below/Above & Left/Right & Big/Small & Tall/Short & Wide/Thin & Behind/Front & Avg. \\
    \midrule
    GPT-4o~\cite{gpt4technicalreport}                                    & --- & 92.11 & 73.33 & 65.38 & 85.71 & 78.57 & 80.47 \\
    GPT-4o mini~\cite{gpt4technicalreport}                               & --- & 68.42 & 66.67 & 30.77 & 52.38 & 53.57 & 54.69 \\
    InternVL2.5-8B~\cite{internvl}                                       & --- & 89.47 & 73.33 & 46.15 & 66.67 & 39.29 & 64.06 \\
    \cellcolor{gray!20}Qwen2.5-VL-7B-Instruct~\cite{qwen2_5}             & --- & 92.11 & 53.33 & 50.00 & 61.90 & 60.71 & 67.19 \\
    \midrule
    SpatialRGPT-VILA-1.5-8B~\cite{cheng2024spatialrgpt}                  & --- & 73.68 & 66.67 & 80.77 & 61.90 & 85.71 & 75.00 \\
    \midrule
    Senna-VLM~\cite{senna}                                               & --- & 21.05 & 6.67 & 11.54 & 14.29 & 28.57 & 17.97 \\
    \cellcolor{gray!20}Cosmos-Reason1-7B~\cite{nvidia2025cosmos-reason1} & --- & 76.32 & 53.33 & 46.15 & 33.33 & 46.43 & 53.91 \\
    \midrule
    \cellcolor{cyan!10}STRIDE-Qwen2.5-VL-7B                             & --- & 81.58 & 66.67 & 61.54 & 71.43 & 60.71 & 69.53 \\
    \cellcolor{cyan!10}STRIDE-Cosmos-Reason1-7B                         & --- & 81.58 & 86.67 & 50.00 & 80.95 & 60.71 & 71.09 \\
    \bottomrule
  \end{tabular*}

  \begin{tabular*}{\linewidth}{@{\extracolsep{\fill}} lcccccc}
    \toprule
    Model & Direct & Horizontal & Vertical & Width & Height & Direction \\
    \midrule
    GPT-4o~\cite{gpt4technicalreport}                                     & 28.21 & 15.00 & --- & 55.56 & 79.31 & 5.88 \\
    GPT-4o mini~\cite{gpt4technicalreport}                                & 7.69  & 17.50 & --- & 55.56 & 82.76 & 14.71 \\
    InternVL2.5-8B~\cite{internvl}                                        & 10.26 & 15.00 & --- & 44.44 & 37.93 & 2.94 \\
    \cellcolor{gray!20}Qwen2.5-VL-7B-Instruct~\cite{qwen2_5}              & 33.33 & 32.50 & --- & 11.11 & 34.48 & 2.94 \\
    \midrule
    SpatialRGPT-VILA-1.5-8B~\cite{cheng2024spatialrgpt}                   & 20.51 & 27.50 & --- & 44.44 & 82.76 & 70.59 \\
    \midrule
    Senna-VLM~\cite{senna}                                                & 2.56  & 0.00  & --- & 0.00  & 0.00  & 0.00 \\
    \cellcolor{gray!20}Cosmos-Reason1-7B~\cite{nvidia2025cosmos-reason1}  & 12.82 & 17.50 & --- & 33.33 & 72.41 & 26.47 \\

    \midrule
    \cellcolor{cyan!10}STRIDE-Qwen2.5-VL-7B                               & 20.51 & 12.50 & --- & 66.67 & 82.76 & 32.35 \\
    \cellcolor{cyan!10}STRIDE-Cosmos-Reason1-7B                           & 12.82 & 5.00  & --- & 66.67 & 79.31 & 17.65 \\
    \bottomrule
  \end{tabular*}

  \vspace{0.2em}
  \begin{tabular*}{\linewidth}{@{\extracolsep{\fill}} lccccccc}
    \multicolumn{8}{l}{\textbf{Object-centric Spatial QA}} \\
    \toprule
    Model & Below/Above & Left/Right & Big/Small & Tall/Short & Wide/Thin & Behind/Front & Avg. \\
    \midrule
    GPT-4o~\cite{gpt4technicalreport}                                    & --- & 44.55 & 36.36 & 37.21 & 31.48 & 42.22 & 39.23 \\
    GPT-4o mini~\cite{gpt4technicalreport}                               & --- & 50.91 & 52.27 & 60.47 & 38.89 & 57.78 & 52.51 \\
    InternVL2.5-8B~\cite{internvl}                                       & --- & 50.91 & 50.00 & 52.33 & 50.00 & 48.89 & 50.74 \\
    \cellcolor{gray!20}Qwen2.5-VL-7B-Instruct~\cite{qwen2_5}             & --- & 61.82 & 61.36 & 50.00 & 40.74 & 53.33 & 54.28 \\
    \midrule
    SpatialRGPT-VILA-1.5-8B~\cite{cheng2024spatialrgpt}                  & --- & 35.45 & 38.64 & 33.72 & 33.33 & 53.33 & 37.46 \\
    \midrule
    Senna-VLM~\cite{senna}                                               & --- & 15.45 & 4.55 & 5.81 & 0.00 & 15.56 & 9.14 \\
    \cellcolor{gray!20}Cosmos-Reason1-7B~\cite{nvidia2025cosmos-reason1} & --- & 54.55 & 45.45 & 40.70 & 35.19 & 66.67 & 48.38 \\
    \midrule
    \cellcolor{cyan!10}STRIDE-Qwen2.5-VL-7B                              & --- & 63.64 & 59.09 & 68.60 & 57.41 & 46.67 & 61.06 \\
    \cellcolor{cyan!10}STRIDE-Cosmos-Reason1-7B                          & --- & 63.64 & 75.00 & 63.95 & 59.26 & 46.67 & 62.24 \\
    \bottomrule
  \end{tabular*}

  \begin{tabular*}{\linewidth}{@{\extracolsep{\fill}} lcccccc}
    \toprule
    Model & Direct & Horizontal & Vertical & Width & Height & Direction \\
    \midrule
    GPT-4o~\cite{gpt4technicalreport}                                     & 20.00 & 16.67 & 13.64 & 55.00 & 63.64 & --- \\
    GPT-4o mini~\cite{gpt4technicalreport}                                & 13.33 & 16.67 & 4.55  & 70.00 & 59.09 & --- \\
    InternVL2.5-8B~\cite{internvl}                                        & 13.33 & 3.33  & 4.55  & 40.00 & 36.36 & --- \\
    \cellcolor{gray!20}Qwen2.5-VL-7B-Instruct~\cite{qwen2_5}              & 6.67  & 3.33  & 0.00  & 5.00  & 18.18 & --- \\
    \midrule
    SpatialRGPT-VILA-1.5-8B~\cite{cheng2024spatialrgpt}                   & 26.67 & 30.00 & 0.00  & 70.00 & 77.27 & --- \\
    \midrule
    Senna-VLM~\cite{senna}                                                & 0.00  & 10.00 & 9.09  & 0.00  & 0.00  & --- \\
    \cellcolor{gray!20}Cosmos-Reason1-7B~\cite{nvidia2025cosmos-reason1}  & 20.00 & 3.33  & 4.55  & 15.00 & 31.82 & --- \\
    \midrule
    \cellcolor{cyan!10}STRIDE-Qwen2.5-VL-7B                               & 40.00 & 53.33 & 40.91 & 75.00 & 95.45 & --- \\
    \cellcolor{cyan!10}STRIDE-Cosmos-Reason1-7B                           & 33.33 & 56.67 & 36.36 & 75.00 & 86.36 & --- \\
    \bottomrule
  \end{tabular*}

  \vspace{0.2em}
  \begin{tabular*}{\linewidth}{@{\extracolsep{\fill}} lccccccc}
    \multicolumn{8}{l}{\textbf{Ego-centric Spatial QA}} \\
    \toprule
    Model & Below/Above & Left/Right & Big/Small & Tall/Short & Wide/Thin & Behind/Front & Avg. \\
    \midrule
    GPT-4o~\cite{gpt4technicalreport}                                   & --- & 36.84 & 43.55 & 28.81 & 50.00 & --- & 37.37 \\
    GPT-4o mini~\cite{gpt4technicalreport}                              & --- & 45.61 & 30.65 & 39.83 & 51.92 & --- & 41.18 \\
    InternVL2.5-8B~\cite{internvl}                                      & --- & 21.05 & 41.94 & 32.20 & 46.15 & --- & 34.60 \\
    \cellcolor{gray!20}Qwen2.5-VL-7B-Instruct~\cite{qwen2_5}            & --- & 49.12 & 30.65 & 45.76 & 44.23 & --- & 42.91 \\
    \midrule
    SpatialRGPT-VILA-1.5-8B~\cite{cheng2024spatialrgpt}                 & --- & 45.61 & 1.61  & 18.64 & 32.69 & --- & 22.84 \\
    \midrule
    Senna-VLM~\cite{senna}                                               & --- & 12.28 & 3.23 & 5.08 & 3.85 & --- & 5.88 \\
    \cellcolor{gray!20}Cosmos-Reason1-7B~\cite{nvidia2025cosmos-reason1} & --- & 29.82 & 30.65 & 29.66 & 34.62 & --- & 30.80 \\
    \midrule
    \cellcolor{cyan!10}STRIDE-Qwen2.5-VL-7B                              & --- & 80.70 & 82.26 & 74.58 & 76.92 & --- & 77.85 \\
    \cellcolor{cyan!10}STRIDE-Cosmos-Reason1-7B                          & --- & 77.19 & 80.65 & 82.20 & 76.92 & --- & 79.93 \\
    \bottomrule
  \end{tabular*}

  \begin{tabular*}{\linewidth}{@{\extracolsep{\fill}} lcccccc}
    \toprule
    Model & Direct & Horizontal & Vertical & Width & Height & Direction \\
    \midrule
    GPT-4o~\cite{gpt4technicalreport}                                   & 25.71 & 3.33  & 13.33 & 68.97 & 33.33 & --- \\
    GPT-4o mini~\cite{gpt4technicalreport}                              & 5.71  & 33.33 & 3.33  & 86.21 & 37.50 & --- \\
    InternVL2.5-8B~\cite{internvl}                                      & 17.14 & 20.00 & 0.00  & 89.66 & 37.50 & --- \\
    \cellcolor{gray!20}Qwen2.5-VL-7B-Instruct~\cite{qwen2_5}            & 20.00 & 0.00  & 0.00  & 17.24 & 20.83 & --- \\
    \midrule
    SpatialRGPT-VILA-1.5-8B~\cite{cheng2024spatialrgpt}                 & 0.00  & 13.33 & 0.00 & 68.97  & 0.00 & --- \\
    \midrule
    Senna-VLM~\cite{senna}                                               & 0.00 & 3.33 & 0.00 & 0.00 & 8.33 & --- \\
    \cellcolor{gray!20}Cosmos-Reason1-7B~\cite{nvidia2025cosmos-reason1} & 14.29 & 0.00 & 6.67 & 75.86 & 20.83 & --- \\
    \midrule
    \cellcolor{cyan!10}STRIDE-Qwen2.5-VL-7B                             & 42.86 & 56.67 & 63.33 & 100 & 100 & --- \\
    \cellcolor{cyan!10}STRIDE-Cosmos-Reason1-7B                         & 51.43 & 40.00 & 63.33 & 100 & 100 & --- \\
    \bottomrule
  \end{tabular*}
  \caption{SpatialRGPT-Bench results. From top to bottom, the table presents original, object-centric QA, and ego-centric QA. All QA categories report success rates \((\uparrow)\).}
  \label{tab:spatialrgpt_details}
\end{table*}

\section{Limitations and Future Work}

\noindent\textbf{Limitations.}\quad
This study has three main limitations. First, because STRIDE-QA is the inaugural benchmark for spatiotemporal reasoning in autonomous driving and no comparable public dataset exists, we cannot yet quantify cross-dataset generalization. Second, resource constraints restricted us to a lightweight LoRA adaptation; the upper-bound performance attainable with full-parameter supervised fine-tuning therefore remains unexplored. Third, although STRIDE-QA is designed to strengthen the spatiotemporal reasoning core of vision–language models, its impact on safety-critical downstream tasks such as motion planning and behavior prediction has not been rigorously assessed.

\noindent\textbf{Future Work.}\quad
The performance of the VLM in predicting, from the ego vehicle’s perspective, the distance, direction, and relative speed of surrounding agents declines when a target object leaves the front camera's FOV. This finding underscores the need for autonomous-driving VLMs to integrate multi-camera inputs and motivates future work on efficiently combining multi-view information. In addition, we will refine the evaluation protocol by adding finer-grained metrics and release it alongside STRIDE-QA so that it can serve as a widely adopted benchmark.

\end{document}